\pdfoutput=1

\documentclass[11pt]{article}
\DeclareUnicodeCharacter{FF1A}{:}
\usepackage{acl}
\usepackage{bbm}
\usepackage{times}
\usepackage{latexsym}
\usepackage{amssymb}   

\usepackage[T1]{fontenc}

\usepackage{makecell}  

\usepackage[utf8]{inputenc}
\usepackage{amssymb}    
\usepackage{colortbl}   
\usepackage{xcolor}     
\usepackage{array}       

\usepackage{hyperref}
\usepackage{url}
\usepackage{booktabs}       
\usepackage{times}  
\usepackage{helvet}  
\usepackage{courier}  
\usepackage{graphicx}
\usepackage{algorithm}
\usepackage{algpseudocode}
\usepackage{amsmath}
\usepackage{multirow}
\usepackage{tabularx}
\usepackage{colortbl}
\usepackage{array}
\usepackage{marvosym}
\usepackage{xcolor}
\usepackage{natbib}  
\usepackage{caption} 
\usepackage{wrapfig}
\usepackage{adjustbox}
\usepackage{lipsum}
\usepackage{soul}
\usepackage{pifont}      
\usepackage{xcolor}      
\definecolor{lightgray}{gray}{0.7}

\usepackage{subfigure}

\newcommand{\warn}[1]{\textcolor{red}{#1}}
\newcommand{\OurDATA}{\textsc{CompKe}}
\newcommand{\OurMODEL}{\textsc{Gdecom-CQA}}
\newcommand{\eat}[1]{}

\newcommand{\keyuan}[1]{\textcolor{orange}{#1}}

\title{\OurDATA{}: Complex Question Answering under Knowledge Editing}

\author{
Keyuan Cheng\textsuperscript{*,1,3,4},
Zijian Kan\textsuperscript{*,1,4}, Zhixian He\textsuperscript{1,5},
Zhuoran Zhang\textsuperscript{1,3}, \\
\textbf{Muhammad Asif Ali\textsuperscript{2},
Ke Xu\textsuperscript{4}, 
Lijie Hu\textsuperscript{†,1,2},
Di Wang\textsuperscript{†,1,2}}\\
$^1$Provable Responsible AI and Data Analytics (PRADA) Lab\\
$^2$King Abdullah University of Science and Technology \\
$^3$Peking University \quad
$^4$South China University of Technology\\
$^5$Sun Yat-sen University 
}

\begin{document}

\maketitle

\begin{abstract}
Knowledge Editing---Efficiently modifying the knowledge in large language models has gathered great attention. Current benchmarks primarily use multi-hop question answering to assess and analyze newly injected or updated knowledge. However, we argue that these benchmarks fail to effectively evaluate how well the updated models apply this knowledge in real-life scenarios, particularly when questions require complex reasoning, involving one-to-many relationships or multi-step logical intersections. To fill in this gap, we introduce a new benchmark, \OurDATA: \underline{\textbf{Comp}}lex Question Answering under \underline{\textbf{K}}nowledge \underline{\textbf{E}}diting, which includes 11,924 complex questions that reflect real-life situations. 
We conduct an extensive evaluation of four knowledge editing methods on~\OurDATA, revealing that their effectiveness varies notably across different models. For instance, MeLLo attains an accuracy of 39.47 on \textsc{GPT-4o-mini}, but this drops sharply to 3.83 on \textsc{Qwen2.5-3B}. We further investigate the underlying causes of these disparities from both methodological and model-specific perspectives. The datasets are available at \url{https://github.com/kzjkzj666/CompKE}.
\end{abstract}

\def\thefootnote{*}\footnotetext{Equal Contribution.}
\def\thefootnote{†}\footnotetext{Corresponding Author.}

\section{Introduction}
\label{sec:intro}
Large language models (LLMs) have demonstrated impressive capabilities across a variety of real-world tasks. However, they are still prone to producing outdated, fraudulent, or incorrect information~\citep{2023_KE_Survey,Zhang2024ACS,fang2023annotations,yao2025understanding,yao2025your,yang2025fraud,su2023detectllm,su2023fake}. To address this, the field of Knowledge Editing (KE)—which focuses on updating a model's knowledge without costly full-model fine-tuning—has emerged as an active area of research~\citep{2023_KE_Survey,Zhang2024ACS}.

\begin{figure}[t]
    \centering
    \includegraphics[width=0.85\linewidth]{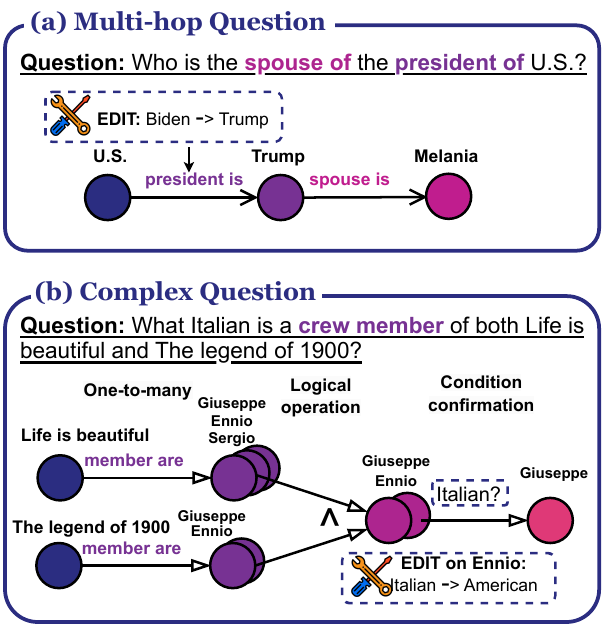}
    \vspace{-1.7ex}
    \caption{(a) An example of a multi-hop question involving only one-to-one 
    sequential step-by-step reasoning. 
    (b) An example of a complex problem involving one-to-many knowledge mapping, logical operations, and conditional confirmation.}
    \label{fig:idea}
    \vspace{-4.1ex}
\end{figure}

A widely adopted strategy for evaluating the effectiveness of KE methods is to test whether the model can reproduce newly injected knowledge, as seen in benchmarks like ZsRE~\citep{levy-etal-2017-zero} and COUNTERFACT~\citep{meng2022locating}. 
We observe, these benchmarks focus mainly on rote memorization and fail to assess whether the model can effectively utilize the updated knowledge in more complex, contextual scenarios.
To address this limitation, MQuAKE~\citep{zhong2023mquake} introduces multi-hop question answering (MQA) as a more rigorous evaluation, requiring models to integrate and reason over multiple pieces of information.
An example in this regard is illustrated in Figure~\ref{fig:idea}, which shows a question: 
\emph{``Who is the spouse of the president of U.S.?''} 
This question requires multiple reasoning steps: 
\textit{(a)} identifying who is the current president of U.S.; and,
\textit{(b)} determining the president's spouse. 

Nevertheless, multi-hop question evaluation remains limited in scope and does not fully capture a model’s ability to flexibly apply newly integrated knowledge. These limitations are evident in three main aspects:
\underline{\textit{(i) Linear question structure}}: The questions typically follow a rigid, sequential pattern, resulting in overly simplistic reasoning chains that can be solved via step by step process.
\underline{\textit{(ii) One-to-one relations}}: Sub-questions are generally based on fact triples with strict one-to-one relationships, which do not reflect the complexity of real-world knowledge. In reality, many facts involve one-to-many relations, such as \textit{``Who are the major shareholders of a company?''}, where a single subject is associated with multiple entities.
\underline{\textit{(iii) Limited edit operations}}: Knowledge edits are mostly restricted to substitutions, neglecting more complex modifications such as additions and deletions that are common in real-world scenarios.


To bridge this gap, we introduce \OurDATA{}: \underline{\textbf{Comp}}lex Question Answering under \underline{\textbf{K}}nowledge \underline{\textbf{E}}diting, a new benchmark specifically designed for complex question answering in the context of knowledge editing. Built from Wikidata,~\OurDATA{} contains 11,924 complex questions. As illustrated in Figure~\ref{fig:idea}(b), \OurDATA{} advances beyond existing multi-hop knowledge editing benchmarks in several key ways:\\
\underline{\textit{\textbf{(i) Diverse question structures}}}: Sub-questions in \OurDATA{} are flexibly composed to form complex questions, incorporating logical operations, conditional checks, and knowledge mapping.\\
\underline{\textit{\textbf{(ii) One-to-many relations}}}: The underlying fact triples support both one-to-one and one-to-many relationships, better reflecting the complexity of real-world knowledge (e.g., questions involving multiple correct answers).\\
\underline{\textit{\textbf{(iii) Expanded capabilities}}}: \OurDATA{} includes a broader range of knowledge edits, systematically covering not only substitutions but also additions and deletions, to more closely mirror real-world knowledge updates.

In order to evaluate the effectiveness of KE methods on \OurDATA{}, we conduct an extensive evaluation of leading KE methods on five LLMs spanning diverse model families, encompassing both open-source and closed-source architectures with a range of parameter sizes. 
Our results reveal that most methods achieve only modest performance on complex question answering tasks. Further analysis across different model scales indicates that parameter-based approaches tend to be more effective for smaller models, while memory-based methods yield better outcomes on larger models with stronger reasoning abilities.
We summarize the key contributions of our work as follows:

\begin{itemize}
\setlength\itemsep{0em}
\item We introduce \OurDATA{}, a novel KE benchmark that overcomes existing 
limitations by incorporating diverse question structures, one-to-many relations, 
and expanded edit types.

\item We comprehensively evaluate major KE methods across five LLMs, uncovering 
significant differences in their ability to handle complex logical problems in 
diverse KE scenarios and providing an in-depth analysis of the underlying 
factors.
\end{itemize}

\eat{\noindent \textit{(ii)} In addition, we solve the above two challenges to some extent and propose a new memory-based editing method \OurMODEL{}: \underline{\textbf{G}}eneric \underline{\textbf{Decom}}position based \underline{\textbf{M}}ulti-hop \underline{\textbf{Q}}uestion \underline{\textbf{A}}nswering. \keyuan{\OurMODEL{} is designed to better adapt to the planning of complex problems \warn{describe how}. It employs a prompt-selection strategy to dynamically construct the most suitable decomposition examples for the problem at hand. \warn{explain the objective} Additionally, it utilizes multi-entity retrieval to ensure high retrieval rates in scenarios such as multiple-choice questions and condition confirmation} \warn{explain what is multi-entity retrieval}}

\section{Related Work}
\label{sec:RW}
\noindent {\bf Knowledge Editing Benchmarks.} 
KE is an essential area of research for LLMs, allowing them to update their knowledge and remain responsive to new or changing information. To evaluate the effectiveness of KE methods, a range of benchmarks have been developed.

Early benchmarks such as COUNTERFACT~\citep{meng2022locating} focus on counterfactual knowledge updates, while ZsRE~\citep{levy-etal-2017-zero} and MzsRE~\citep{wang2023retrievalaugmentedmultil} expand evaluation to zero-shot and multilingual scenarios. ECBD~\citep{onoe2023lmslearnnewentities} investigates whether newly injected facts can support reasoning over related entities. EasyEdit~\citep{Wang2023EasyEditAE} provides a unified framework for implementing and comparing various state-of-the-art KE approaches. More recent efforts, including MQuAKE~\citep{zhong2023mquake} and MQA-AEVAL~\citep{ali2024mqakealmultihop}, extend KE evaluation to multi-hop reasoning tasks. TEMPLAMA~\citep{zheng-etal-2023-edit} and ATOKE~\citep{yin2023historymatterstemporalknowledge} address temporal knowledge editing, aiming to update time-sensitive information without interfering with knowledge from other periods.

Despite these advances, existing benchmarks often fail to capture the full complexity of real-world scenarios. 
In particular, they typically lack support for reasoning over one-to-many relations and for combining entities 
using logical operations such as intersection and union.

\noindent {\bf Knowledge Editing Methods.}
Existing research on KE can be broadly categorized into parameter-based and memory-based approaches.

\noindent \textit{Parameter-based methods} update a model’s internal parameters to encode new or corrected knowledge. Notable examples include ROME~\citep{meng2022locating} and MEMIT~\citep{meng2022mass}, which target and modify parameters linked to specific facts, and Transformer-Patcher~\citep{huang2023transformer}, which introduces new neurons to encode edits. To address issues such as high computational cost and catastrophic forgetting, lightweight adaptation techniques like LoRA~\citep{hu2021loralowrankadaptationlarge}, Prompt Tuning~\citep{shi2024deptdecomposedprompttuning}, and QLoRA~\citep{dettmers2023qloraefficientfinetuningquantized} have been proposed.
Despite their effectiveness for single-fact edits, parameter-based methods often struggle with multi-hop and complex reasoning tasks. Additionally, they are not applicable to closed-source models (e.g., OpenAI GPTs) that are only accessible via APIs, and they generally require more computational resources compared to memory-based approaches.

\noindent \textit{Memory-based methods} maintain an external memory to store knowledge edits, retrieving relevant information at inference time~\cite{cheng2024multi}. For example, SERAC~\citep{Mitchell2022MemoryBasedME} combines semi-parametric editing with retrieval-augmented models, while GRACE~\citep{Hartvigsen2022AgingWG} leverages adapters and vector matching for knowledge modification. IKE~\citep{Zheng2023CanWE} uses in-context learning with stored demonstrations, and MeLLo~\citep{zhong2023mquake} stores edited facts externally, incorporating them via prompts. PokeMQA~\citep{gu2023pokemqa} introduces a two-stage process for question decomposition and conflict detection, and GLAME~\citep{zhang2024knowledgegraphenhancedlarge} integrates a knowledge graph module to improve retrieval.

In our analysis, we find that MeLLo and PokeMQA are particularly effective for multi-hop reasoning. Consequently, we adopt them as baselines in our experiments to evaluate the generalization of memory-based methods to complex question answering. Additional discussion of related work is provided in Appendix~\ref{app:related}.

\vspace{-1.7ex}
\section{Preliminaries}
\vspace{-1.7ex}
\label{sec:background}
\noindent{\bf Notations.} We represent the knowledge base as a set of triples $\mathcal{D} = \{(s, r, o)\} \subseteq \mathcal{E} \times \mathcal{R} \times \mathcal{E}$, where $\mathcal{E}$ is the set of entities and $\mathcal{R}$ is the set of relations. Each triple $(s, r, o)$ encodes a factual statement indicating that the subject entity $s$ is connected to the object entity $o$ via relation $r$. To accommodate one-to-many relationships, we generalize the knowledge instance to the form $(s, r, \mathcal{O})$, where $\mathcal{O} = \{o_1, o_2, \ldots\}$ is a set of object entities. For example, (Avatar, actors$\_$are, $\{$Worthington, Saldana, $\ldots\}$) captures that the movie Avatar has multiple actors.

\eat{We define that if $(s,r,o)\in \mathcal{D}$, then there exists a \textit{relation mapping} between $s$ and $o$ under the knowledge base $\mathcal{D}$.}

\subsection{Complex Questions}
\label{sec:complex_questions}
\eat{\st{Previous work has not provided a systematic definition of 
complex questions, so in this section we will introduce our definition of complex questions.}}

\noindent
Building on the example introduced earlier, we now formally define the notion of complex questions considered in this work. For a brief overview of multi-hop question answering (MQA) and MQA in the context of knowledge editing (KE), please refer to Appendix~\ref{app:mqa}.
\begin{figure}
    \centering
    \includegraphics[width=0.99\linewidth]{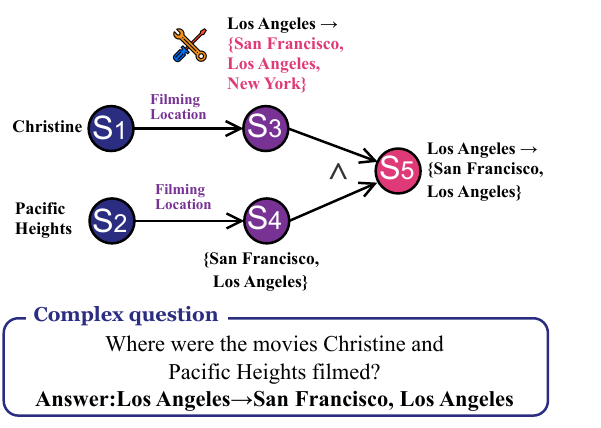}
    \vspace{-2.7ex}
    \caption{An example of a complex question under knowledge editing. Knowledge editing 
    occurs in the first sub-question, where the filming location of \textit{Christime} 
    is modified from \{ Los Angeles\} to \{San Francisco, Los Angeles, New York\}.}
    \vspace{-2.7ex}
    \label{fig:datasets}
\end{figure}

We represent a complex question $Q$ as a graph-structured reasoning problem, that is,
$Q = (\mathbf{S}, \mathbf{L})$,
where $\mathbf{S} = \{S_1, S_2, \ldots\}$ is a collection of \textit{intermediate entity sets}, and $\mathbf{L} = \{L_1, L_2, \ldots\}$ is a collection of \textit{reasoning links}.
Each $S_i \in \mathbf{S}$ is itself a set of entities, i.e., $S_i = \{s_1, s_2, \ldots\}$, which allows us to naturally capture both one-to-one and one-to-many knowledge relations.
Each $L_i \in \mathbf{L}$ denotes a reasoning link. Unlike standard relations in knowledge graphs—which simply connect one entity $s_i$ to another $s_j$ via a relation $r$—our reasoning links generalize this notion to support richer operations, including conditional confirmation and logical operations, as detailed below.

\paragraph{Reasoning Links.}
We categorize the reasoning links into two distinct categories:

\definecolor{c1}{RGB}{255,200,200}
\definecolor{c2}{RGB}{150,200,255} 
\definecolor{c3}{RGB}{150,200,255}

\paragraph{\underline{\textit{(a) \bf Knowledge-related Links:}}}
These links enable traversal between sets of entities, allowing the reasoning process to progress from one set $S_i \in \mathbf{S}$ to another set $S_j \in \mathbf{S}$ based on the underlying knowledge base. We further distinguish the following types:

\noindent {\textit{(i) Knowledge Mapping.}} 
Given a set $S_i$, a knowledge mapping link connects $S_i$ to the set of adjacent entities $S_j = \bigcup_{s \in S_i} A_r(s)$, where $A_r(s) = \{ s' \mid (s, r, s') \in \mathcal{D} \}$ denotes all entities $s'$ related to $s$ via relation $r$.

\noindent {\textit{(ii) Condition Confirmation.}} 
Given a relation $r$ and a target entity $s'$, this link selects the subset of entities from $S_i$ that satisfy the condition of being connected to $s'$ via $r$. Formally, $S_j = \{ s \in S_i \mid (s, r, s') \in \mathcal{D} \}$, i.e., all $s$ in $S_i$ for which the triple $(s, r, s')$ exists in the knowledge base. This operation checks whether each $s$ in $S_i$ is related to $s'$ through $r$.

\eat{
For instance, for the question:
\textit{``Whether Giuseppe and Ennio are Italian nationals?''}, $S_i$ 
are $\{\textit{Giueppe},\textit{Ennio}\}$, $s^\prime$ is \{Italian\} 
and $r$ is \{national$\_$of\}.

\warn{what is $s$ here..?}

\warn{$C_1(Giueppe,national~of,Italian)$
 $C_2(Ennio,national~of,Italian)$, means check if Giue... is Italian, and check if ennio is Italian.}
}

\paragraph{\underline{\textit{(b) \bf Logical Links:}}} 
Logical links enable the application of set-based logical operations over collections of intermediate entity sets $\{ S_{1}, S_{2}, \ldots, S_{n} \} \subseteq \mathbf{S}$. These operations facilitate more expressive reasoning by combining or filtering entities across multiple sets. The primary logical operations considered are:

\noindent \textit{(i) Intersection.} The intersection operation returns the set of entities that are present in all input sets, formally defined as $S_j = \bigcap_{k=1}^{n} S_k$.

\noindent \textit{(ii) Union.} The union operation aggregates all entities that appear in any of the input sets, given by $S_j = \bigcup_{k=1}^{n} S_k$.

\paragraph{\bf Example 1.} Figure~\ref{fig:datasets} provides an illustrative example of a complex question involving multiple reasoning links: \textit{``Where were the movies Christine and Pacific Heights filmed?''}. 
The intermediate entity sets are as follows: 
$S_1 = \{\text{Christine}\}$, 
$S_2 = \{\text{Pacific Heights}\}$, 
$S_3 = \{\text{Los Angeles}\}$, 
$S_4 = \{\text{San Francisco}, \text{Los Angeles}\}$, and 
$S_5 = \{\text{Los Angeles}\}$. 
The reasoning proceeds in three steps: 
(1) $L_1$: map $S_1$ to $S_3$ via the \texttt{filming\_at} relation; 
(2) $L_2$: map $S_2$ to $S_4$ via the same relation; 
(3) $L_3$: apply a logical intersection between $S_3$ and $S_4$ ($S_3 \cap S_4$) 
to yield the final answer, $S_5 = \{\text{Los Angeles}\}$.

\paragraph{\bf Complex Question Answering under Knowledge Editing.}
We formalize a knowledge edit as $e = (s, r, \mathcal{O} \to \mathcal{O}^\prime)$, where the object set $\mathcal{O}$ associated with subject $s$ and relation $r$ is updated to a new set $\mathcal{O}^\prime$, supporting one-to-many modifications. The model is assumed to have access to the original knowledge base $\mathcal{D}$. Given a set of edits $\mathcal{E} = \{e_1, e_2, \ldots\}$, we define the set of knowledge to be removed as $\mathcal{D}_{del}^\mathcal{E} = \{(s_i, r_i, \mathcal{O}_i) \mid e_i \in \mathcal{E}\}$ and the set of knowledge to be added as $\mathcal{D}_{add}^\mathcal{E} = \{(s_i, r_i, \mathcal{O}_i^\prime) \mid e_i \in \mathcal{E}\}$. The updated knowledge base is then given by:
$\mathcal{D}^\prime = (\mathcal{D} - \mathcal{D}_{del}^\mathcal{E}) \cup \mathcal{D}_{add}^\mathcal{E}$.
The objective is that, following the application of edits and the resulting update of the knowledge base to $\mathcal{D}^\prime$, the LLM is able to correctly answer the complex question $Q$ by utilizing the modified knowledge.

\vspace{-1.7ex}
\section{\OurDATA{}}
\vspace{-1.7ex}
\label{sec:data_curate}
\begin{figure*}[t]
    \centering
    \includegraphics[width=0.99\linewidth]{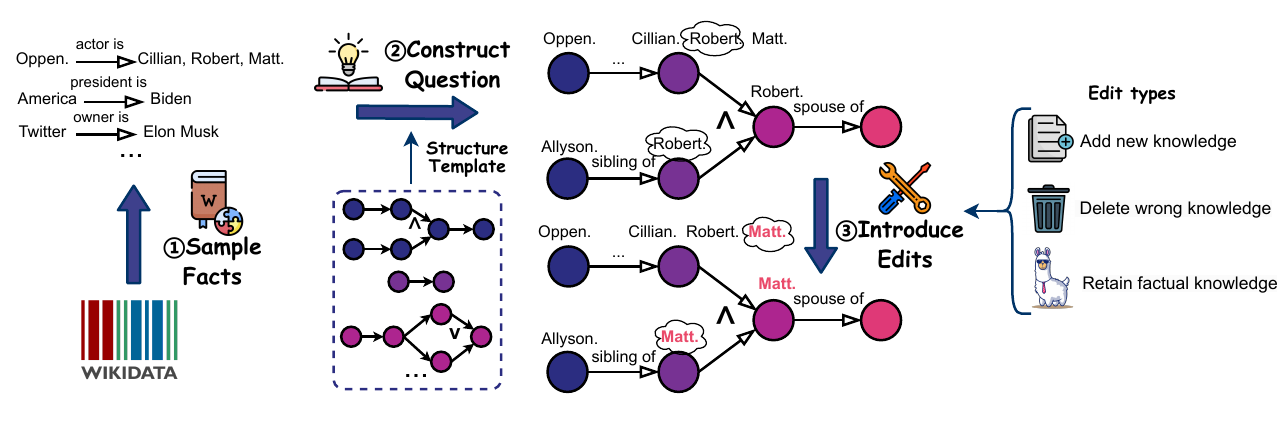}
    \vspace{-1.7ex}
    \caption{The construction process of \OurDATA{}.}
    \label{fig:dataprocess}
    \vspace{-3.7ex}
\end{figure*}

\definecolor{c3}{RGB}{223, 57, 119} 
While complex questions frequently arise in real-world scenarios, they are insufficiently addressed in LLM-based question answering within the context of knowledge editing. Existing benchmarks mainly emphasize linear multi-hop questions, which constrains their ability to assess more intricate queries. To address this limitation, we introduce \OurDATA{}: \underline{\textbf{Comp}}lex Question Answering under \underline{\textbf{K}}nowledge \underline{\textbf{E}}diting. Below, we present an overview of~\OurDATA{} and outline the key steps of its construction process.

\eat{Compared with multi-hop questions, the difficulty of constructing 
complex questions is: 1) The reasoning structure of complex problems 
has complex graph features, while multi-hop problems have simple linear 
reasoning structures; 2) How to ensure the practical relevance of complex 
problems while avoiding inappropriate or irrelevant questions.}
\subsection{Dataset Construction}

\paragraph{\bf Overview.}
The workflow of data construction process, illustrated in Figure~\ref{fig:dataprocess}, consists of six main steps. We begin by extracting factual triples from Wikidata. In the second step, we select relevant relations and sample corresponding triples. Third, we assemble these triples into complex questions featuring diverse reasoning structures, introducing edits at suitable points within the questions. Step four involves applying counterfactual modifications, followed by step five, where we filter out conflicting instances to ensure consistency. In the final step, the structured questions are transformed into natural language. Each step is described in more detail below.

\paragraph{\bf Step 1: Collecting Relation Templates.}
We begin by selecting 37 relations from Wikidata’s \textit{List of Properties} using a two-stage approach. First, we focus on one-to-many relations (such as family-child, book-authors, and movie-actors), which are crucial for mapping knowledge that connects a single entity to multiple others. Second, we add one-to-one relations (such as country-capital and person-hometown) that represent core attributes of entities, supporting both direct mapping and conditional reasoning. To ensure the dataset’s real-world relevance, we give preference to relations that are frequently encountered in everyday contexts. The complete set of relation templates used in \OurDATA{} is listed in Appendix Table~\ref{Relation}.

\paragraph{\bf Step 2: Sampling Facts.} 
After selecting the relation templates, we construct the knowledge base $\mathcal{D}$, prioritizing widely recognized facts over obscure ones. We sample single-hop knowledge triples from Wikidata based on the chosen relation templates and rank them according to their frequency of access, giving precedence to the most commonly referenced triples. To further ensure the relevance and recallability of the knowledge, we utilize \textsc{GPT-3.5-Turbo-instruct} to filter out facts that the model cannot recall. The resulting knowledge base $\mathcal{D}$ forms the foundation for generating complex questions.

\eat{After relation templates, we need 
to construct knowledge base $\mathcal{D}$. 
For this, we want the knowledge to be included in the questions
to be relatively common rather than obscure. Based on the collected
relation templates, we sample single-hop knowledge triples from
Wikidata and rank them according
to their frequency of access, with the most frequently accessed triples
placed at the top. We then use GPT-3.5-Turbo to filter out knowledge
that the model cannot recall. We will use this collected knowledge
$\mathcal{D}$ to create complex questions.}

\paragraph{\bf Step 3: Constructing Complex Questions.}
Complex questions typically exhibit structured reasoning patterns, as illustrated in Figure~\ref{fig:datasets}, where knowledge mapping is often combined with logical operations such as intersection. To systematically capture these reasoning structures, we begin by manually curating a set of high-quality complex questions to serve as seed examples. From these, we abstract their underlying reasoning structures by removing intermediate entities, resulting in reusable templates. These templates are then instantiated with real-world facts from $\mathcal{D}$ to generate concrete complex questions. The instantiation process starts by randomly selecting the leaf nodes, and then iteratively determining the intermediate entities through logical operations or by referencing knowledge in $\mathcal{D}$, repeating this process until all entities in the reasoning structure are specified.

To maintain the quality and relevance of the generated questions, we apply several filtering criteria during instantiation: 
\textit{(i)} questions that lack a valid answer,
\textit{(ii)} questions where the set of intermediate entities is empty, and
\textit{(iii)} questions in which entities participating in logical operations are of incompatible types. 
Representative examples of relational structures and their instantiated complex questions are shown in the Appendix (Figure~\ref{Structure}).

\eat{We observe that complex questions may be organized as 
a reasoning structures, such as the example in 
Figure~\ref{fig:datasets}, which first undergoes 
knowledge mapping and followed by logical 
operations, \textit{e.g.,} intersection. To collect these common reasoning 
structures, we start by manually constructing a subset of high-quality complex 
questions to act as seed. Next, we remove the intermediate entities from these questions to extract 
the underlying reasoning structure. We use these reasoning structures a template to generate specific complex questions by instantiating it with real-world facts from $\mathcal{D}$. The process proceeds as follows: we randomly initialize the leaf nodes of the reasoning structure. From there, we use logical operation or knowledge in $\mathcal{D}$ to progressively identify the intermediate entities at next step. This process is repeated iteratively until all entities, including the intermediate ones, are fully determined.
To ensure the practical relevance of the instantiated questions, 
we filter out cases exhibiting the following conditions:
\textit{(i)} questions with no answer;
\textit{(ii)} questions that result in an empty set of intermediate entities;
\textit{(iii)} cases where the entities involved in the logical 
operations are of different types, making the logic incompatible
\textit{etc.} 
For illustration, we show some exemplar relational structures 
along with complex questions instantiated from them in the 
Appendix (Figure \ref{Structure}). 
}
\paragraph{\bf Step 4: Introducing Counterfactual Edits.}
To simulate real-world knowledge updates, we introduce counterfactual edits into the knowledge base. For each complex question, we randomly select a knowledge mapping of the form $(s, r, \mathcal{O})$ and apply an edit $e = (s, r, \mathcal{O}^\prime)$. In contrast to prior benchmarks that focus solely on one-to-one relations and simple entity substitutions, our approach supports edits on one-to-many relations, resulting in more intricate modifications. To systematically capture the nature of these changes, we define three fundamental operations—addition, deletion, and retention—which can be combined to represent any edit:

\noindent \underline{\textit{(i)} Addition:} 
$\mathcal{O}_{\text{add}}=\mathcal{O}^\prime \setminus \mathcal{O}$, 
where $\mathcal{O}_{\text{add}}$ represents the set of newly added entities;

\noindent \underline{\textit{(ii)} Deletion:} 
$\mathcal{O}_{\text{del}}=\mathcal{O}\setminus \mathcal{O}^\prime$, 
where $\mathcal{O}_{\text{del}}$ represents the set of removed entities;

\noindent \underline{\textit{(iii)} Retention:} 
$\mathcal{O}_{\text{ret}}=\mathcal{O}\cap\mathcal{O}^\prime$, where 
$\mathcal{O}_{\text{ret}}$ represents the set of retained entities.

\paragraph {\bf Example 2.}
We provide an example of a counterfactual edit involving the management of ``Microsoft'': 
(Microsoft, managers$\_$are, \{John, Smith, Dave\} $\to$ \{Smith, Eden, Keyes\}). 
In this case, the edit deletes \{John\}, retains \{Smith\}, and adds \{Eden, Keyes\}.

\begin{table}[t]
    \centering
    \resizebox{1.0\columnwidth}{!}{
    \begin{tabular}{cl}
    \toprule
    $\mathcal{E}$ & (\textit{Christine}, filming location, \{ Los Angeles\}$\to$ \\
    &\{{San Francisco, Los Angeles, New York}\}) \\
    \midrule
    $\mathcal{Q}$ 
    &\textit{i)Where were the movies Christine and Pacific Heights filmed?}\\
    &\textit{ii)}In which locations were both the movie \textit{Christine} and \\&\textit{Pacific Heights} filmed?\\
    &\textit{iii)}What were the filming locations for both the movie \textit{Christine} \\&  and \textit{Pacific Heights}?\\
    \midrule
    $\mathcal{A}$& \{Los Angeles\}\\
    $\mathcal{A}^*$ &\{San Francisco,
Los Angeles\} \\
    \midrule
    $\mathcal{T}$ &(\textit{Christine}, filming location,\{Los Angeles\})\\
    &(\textit{Pacific
Heights}, filming location,\{San Francisco,
Los Angeles\})\\
    $\mathcal{T}^*$  &(\textit{Christine}, filming location,\{{San Francisco, Los Angeles, New York}\})\\
     &(\textit{Pacific
Heights}, filming location,\{San Francisco,
Los Angeles\})\\
  \midrule
$\mathcal{L}$ & $\{ \text{Los Angeles} \} \cap \{ \text{San Francisco}, \text{Los Angeles} \} = \{ \text{Los Angeles} \}$ \\
$\mathcal{L}^*$ & $\{ \text{San Francisco}, \text{Los Angeles}, \text{New York} \} \cap \{ \text{San Francisco}, \text{Los Angeles} \} $\\
&=$ \{ \text{San Francisco}, \text{Los Angeles} \}$
\\
    \bottomrule
    \end{tabular}
    }
\caption{A case from \OurDATA{}, illustrating the components involved in question editing. Here, $\mathcal{E}$ represents the edit, $\mathcal{Q}$ is the natural language question, $\mathcal{A}$ and $\mathcal{A}^*$ denote the answers before and after editing respectively. $\mathcal{T}$ and $\mathcal{T}^*$ are the sets of fact triples before and after editing, which form the complex question. Additionally, $\mathcal{L}$ and $\mathcal{L}^*$ indicate the logic operations applied to the question before and after editing.}
    \label{tab:data_case_example}
\vspace{-2.1ex}
\end{table}

\paragraph{\bf Step 5:  Filtering Conflicting Edits.}
Since the counterfactual edits in Step 4 are introduced randomly, for a 
batch of edits  $\mathcal{E}= \{e_1, e_2, \dots\}$ there may be edits 
corresponding to different cases where 
\( e_i = (s_i, r_i, \mathcal{O}_i \to \mathcal{O}_i^*) \) and 
\( e_j = (s_j, r_j, \mathcal{O}_j \to \mathcal{O}_j^*) \), with 
\( s_i = s_j \) and \( r_i = r_j \), but 
\( \mathcal{O}_i^* \neq \mathcal{O}_j^* \). 
This implies that conflicting facts may exist within the same batch, which can undermine the validity of the evaluation if introduced together. To address this, we detect and group all conflicting cases, and then randomly retain only one instance from each group.

\eat{This indicates that two contradictory new facts have been introduced in the batch. If such conflicting facts are fed into the model in the same batch, it would severely undermine the evaluation's validity. To address this, we iterate through each case, group the conflicting ones and randomly select one to retain.}

\paragraph{\bf Step 6: Phrasing in Natural Language.}
Building on steps 1–5, we generate complex questions involving edits, where each question consists of multiple fact triples. To enable evaluation by LLMs, these structured questions are converted into natural language. Specifically, for each reasoning structure defined in Step 3, we manually curate eight high-quality examples. Using GPT-4o-mini, we then generate three natural language variants for each structured question. Additional details on construction can be found in Appendix~\ref{app:our_data}.

\eat{Building on steps 1-5, we derived complex questions involving edits, each composed of multiple fact triples. To facilitate the evaluation by the large model, these questions must be converted into natural language. Specifically, for each reasoning structure defined in Step 3, we first manually curate eight high-quality examples. Then, using GPT-4o-mini, we generate three natural language questions for each structured question.}

\begin{table}[b!]
\vspace{-1.7ex}
\centering
\resizebox{1.0\linewidth}{!}{
\begin{tabular}{lcccccc}
\toprule
\textbf{\#Edits} & \textbf{1} & \textbf{2} & \textbf{3} & \textbf{4} & \textbf{5} & \textbf{Total} \\
\midrule
\texttt{Edit\_num} & 9,697 & 1,118  &998 &103   & 8  & 11,924 \\ \texttt{Step\_num} & 200 & 424  & 5,770  & 2,949   & 2,581  & 11,924  \\
\bottomrule
\end{tabular}}
\caption{Statistical Results of \OurDATA{}.}
\label{tab:stat_results}
\vspace{-1.7ex}
\end{table}

\subsection{Dataset Summary}
Table~\ref{tab:stat_results} summarizes the distribution of our dataset along two key axes: \texttt{Edit\_num} and \texttt{Step\_num}. Here, \texttt{Edit\_num} indicates how many triples are edited within each complex question. The majority of questions in \OurDATA{} involve a single edit, while a smaller proportion feature two or more edits. \texttt{Step\_num} captures the number of reasoning steps required to solve each question. Most questions require three reasoning steps, with four-step and five-step questions appearing less frequently. This distribution highlights the predominance of moderately complex questions in our dataset, while still providing a range of multi-step and multi-edit scenarios for comprehensive evaluation.
\eat{Table~\ref{tab:stat_results} presents the dataset distribution across two 
dimensions: \texttt{Edit\_num} and \texttt{Step\_num}. \texttt{Edit\_num} refers 
to the number of triples edited in a complex question. 
In \OurDATA{}, the majority of cases involve the editing of a single triple, 
with two-edited cases ranking second. \texttt{Step\_num} indicates the number of 
reasoning steps required to solve a complex question, with 3-step questions 
being the most common, followed by 4-step questions.}

\paragraph{\bf Example 3.}
Table~\ref{tab:data_case_example} provides a representative example from \OurDATA{}, showcasing a complex question formed by merging two sub-questions using an intersection operation. In this example, the edit occurs in the first sub-question, where Christine's filming locations are modified from \{Los Angeles\} to \{San Francisco, Los Angeles, New York\}. As a result, San Francisco is included in the final answer.

\vspace{-1.7ex}
\section{Experiments}
\vspace{-1.7ex}
\begin{table*}[t]
\centering
\resizebox{0.99\linewidth}{!}{
\begin{tabular}{ccccccccccc}
\toprule[1.0pt]
\multirow{2}{*}{\textbf{Model}} & \multirow{2}{*}{\textbf{Method}} & \multicolumn{3}{c}{\textbf{1-edited}} & \multicolumn{3}{c}{\textbf{100-edited}} & \multicolumn{3}{c}{\textbf{3000-edited}} \\ \cmidrule{3-11} 
& & \texttt{Aug}  & \texttt{Ret} & \texttt{Acc} & \texttt{Aug}  & \texttt{Ret} & \texttt{Acc}  & \texttt{Aug}  & \texttt{Ret}  & \texttt{Acc} \\ 
\hline
\multirow{4}{*}{\scshape \textbf{Qwen2.5-3B}}  
& ROME    & 12.61  & 17.91 & 15.26 & 4.80 & 4.40 & 4.60  & 0.82  &1.59& 1.21 \\ 
& MEMIT   &  \textbf{20.99}& \textbf{23.86} & \textbf{22.43} & \textbf{7.80} & \textbf{6.73} & \textbf{7.27}  & \textbf{1.52} & \textbf{3.75} & \textbf{2.64}\\
& MeLLo   & 5.40 & 2.25 & 3.83 & 3.06    & 3.39& 3.23 & 0.69 & 2.00& 1.35\\
& PoKeMQA & 4.26  & 1.85 & 3.06 & 2.85  &1.38 & 2.12 & 0.71 & 0.62& 0.67  \\
\hline
\multirow{4}{*}{\scshape \textbf{Qwen2.5-7B}}  
& ROME    &22.82  & 25.09 & 23.96 & 7.50 & 7.98  & 7.74 & 0.73  & 0.98& 0.86\\ 
& MEMIT   & \textbf{29.40} & \textbf{27.72} & \textbf{28.56} & \textbf{24.11} & \textbf{24.80}  & \textbf{24.46} & 1.88 & 2.05 & 1.97\\
& MeLLo   & 17.78  & 13.38& 15.58 &10.35 & 17.32  & 13.84 & \textbf{8.98} & \textbf{12.59}& \textbf{10.79 }\\
& PoKeMQA & 15.59  & 11.41 & 13.50 &  8.17 & 13.67& 10.92 & 5.04 & 9.15 & 7.10\\
\hline
\multirow{4}{*}{\scshape \textbf{LLaMa-3.1-8B}}  
& ROME    & 7.44  & 24.84& 16.14 & 1.50 &1.14   & 1.32 & 0.56  & 0.61& 0.59\\ 
& MEMIT   & 4.90 & \textbf{33.22}& \textbf{19.06} & 5.00 & \textbf{29.27} & \textbf{17.14}  & 5.03 & \textbf{29.20}& \textbf{17.12} \\
& MeLLo   & \textbf{14.06}  & 17.95& 16.00 & \textbf{9.17} & 17.84  & 13.51 & \textbf{8.98} & 14.17 & 11.58\\
& PoKeMQA & 11.40  & 15.10& 13.25 & 8.87  & 16.85 & 12.86 & 7.45& 12.73 & 10.09\\
\hline
\multirow{2}{*}{\scshape \textbf{GPT-3.5-turbo}}  
& MeLLo   & \textbf{49.21}  & \textbf{44.88}& \textbf{47.05} & \textbf{37.10} &\textbf{44.09} & \textbf{40.60} &\textbf{32.61}  &  \textbf{38.58}& \textbf{35.60}\\
& PoKeMQA & 23.20  & 25.15 & 24.18 & 21.47 & 23.28 & 22.38 & 20.20 & 22.20 & 21.20\\
\hline
\multirow{2}{*}{\scshape \textbf{GPT-4o-mini}}  
& MeLLo   & 22.07 & 25.19& 23.63 & 20.31 & 23.62& 21.96  & 18.75 & 22.14 & 20.45  \\
& PoKeMQA & \textbf{36.60} & \textbf{42.33}& \textbf{39.47} & \textbf{35.42}  & \textbf{41.35}& \textbf{38.39}  & \textbf{28.36}& \textbf{35.02} & \textbf{31.69} \\
\bottomrule[1.0pt]
\end{tabular} }
\caption{Experimental results for~\OurDATA{}. We~\textbf{boldface} the best results.}
\label{tab:main_results}
\vspace{-2.1ex}
\end{table*}

\begin{figure*}[htbp]
    \centering
    \begin{minipage}[b]{0.32\linewidth}
        \includegraphics[width=\linewidth]{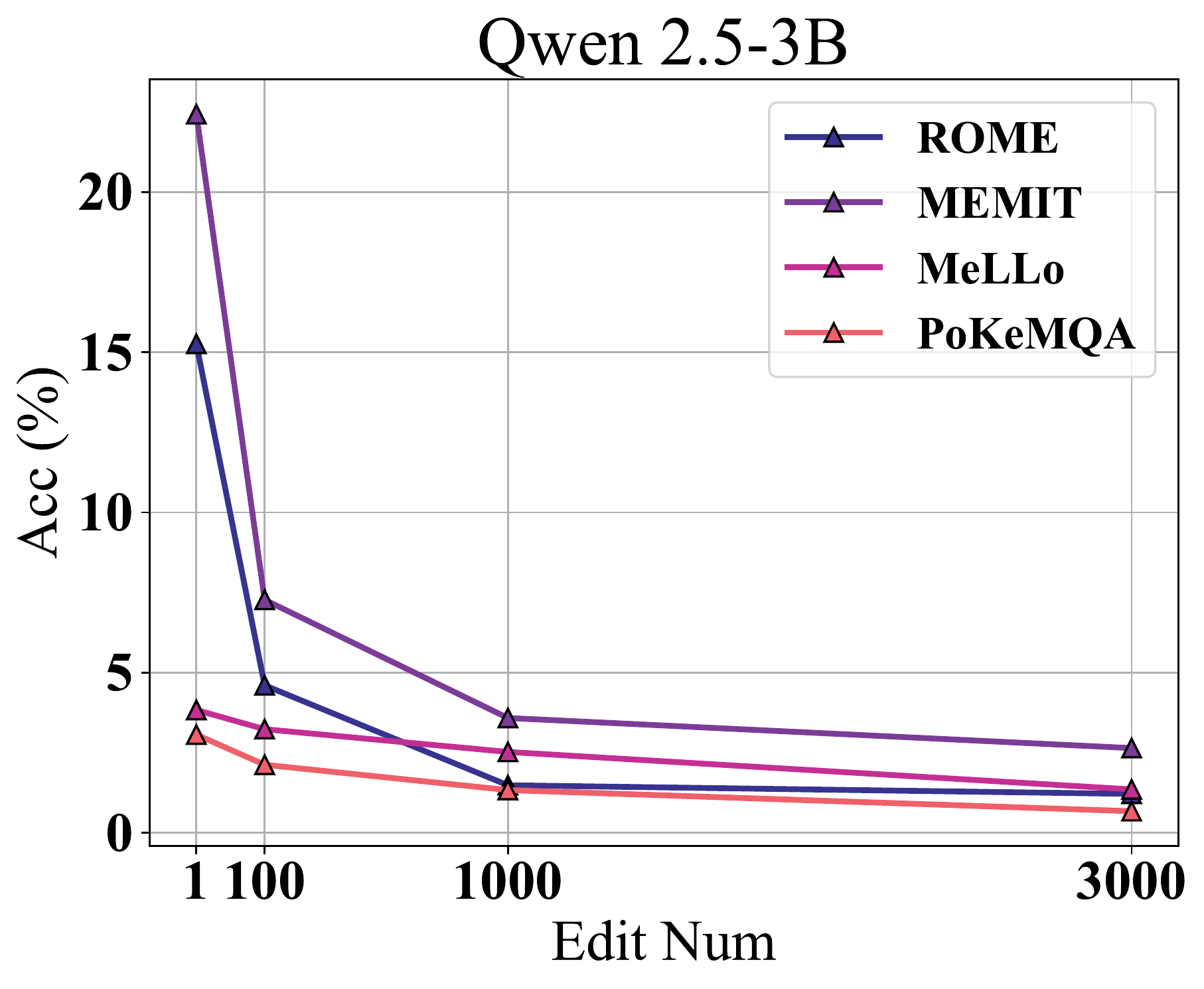}
    \end{minipage}
    \hfill
    \begin{minipage}[b]{0.32\linewidth}
        \includegraphics[width=\linewidth]{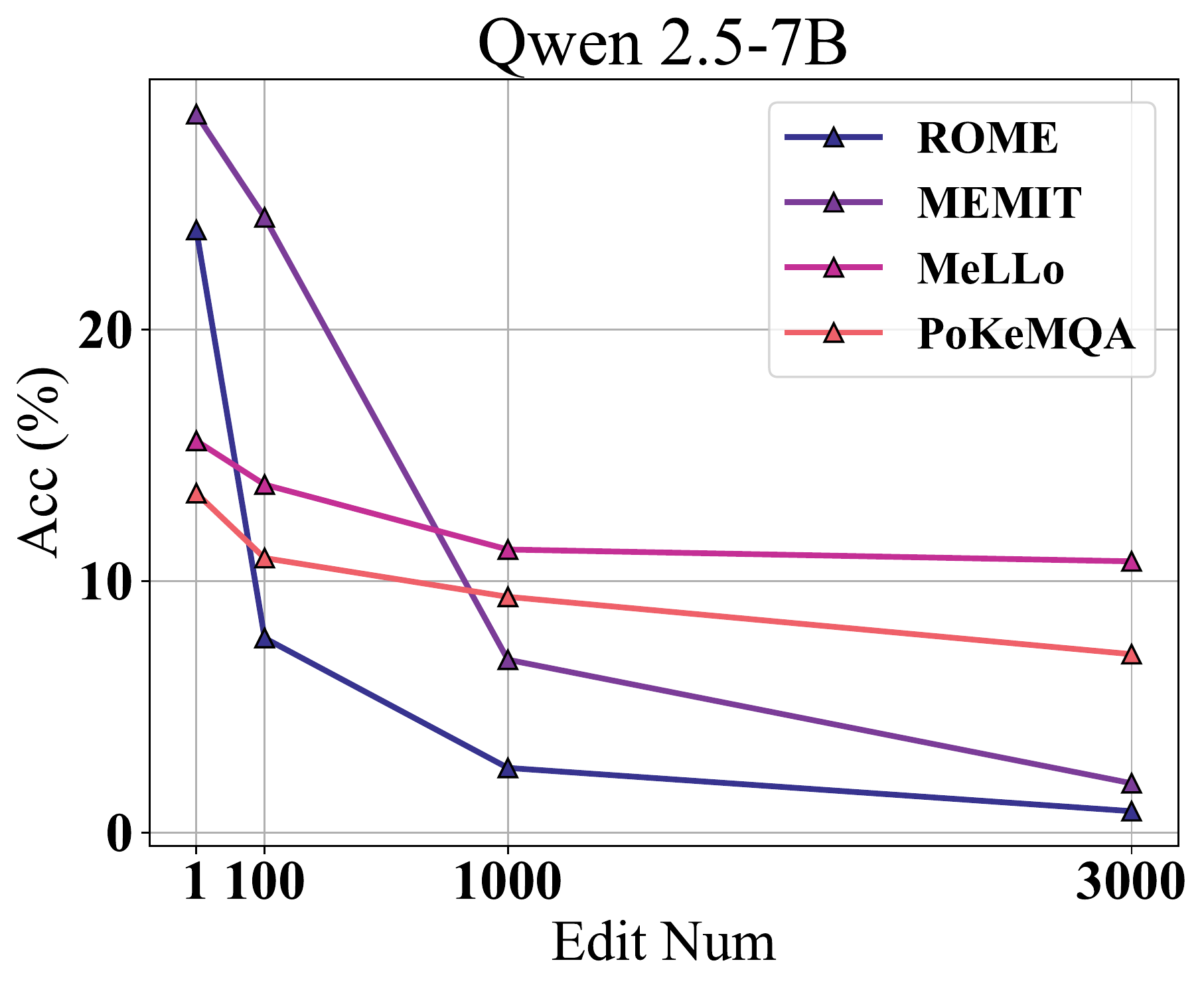}
    \end{minipage}
    \hfill
    \begin{minipage}[b]{0.32\linewidth}
        \includegraphics[width=\linewidth]{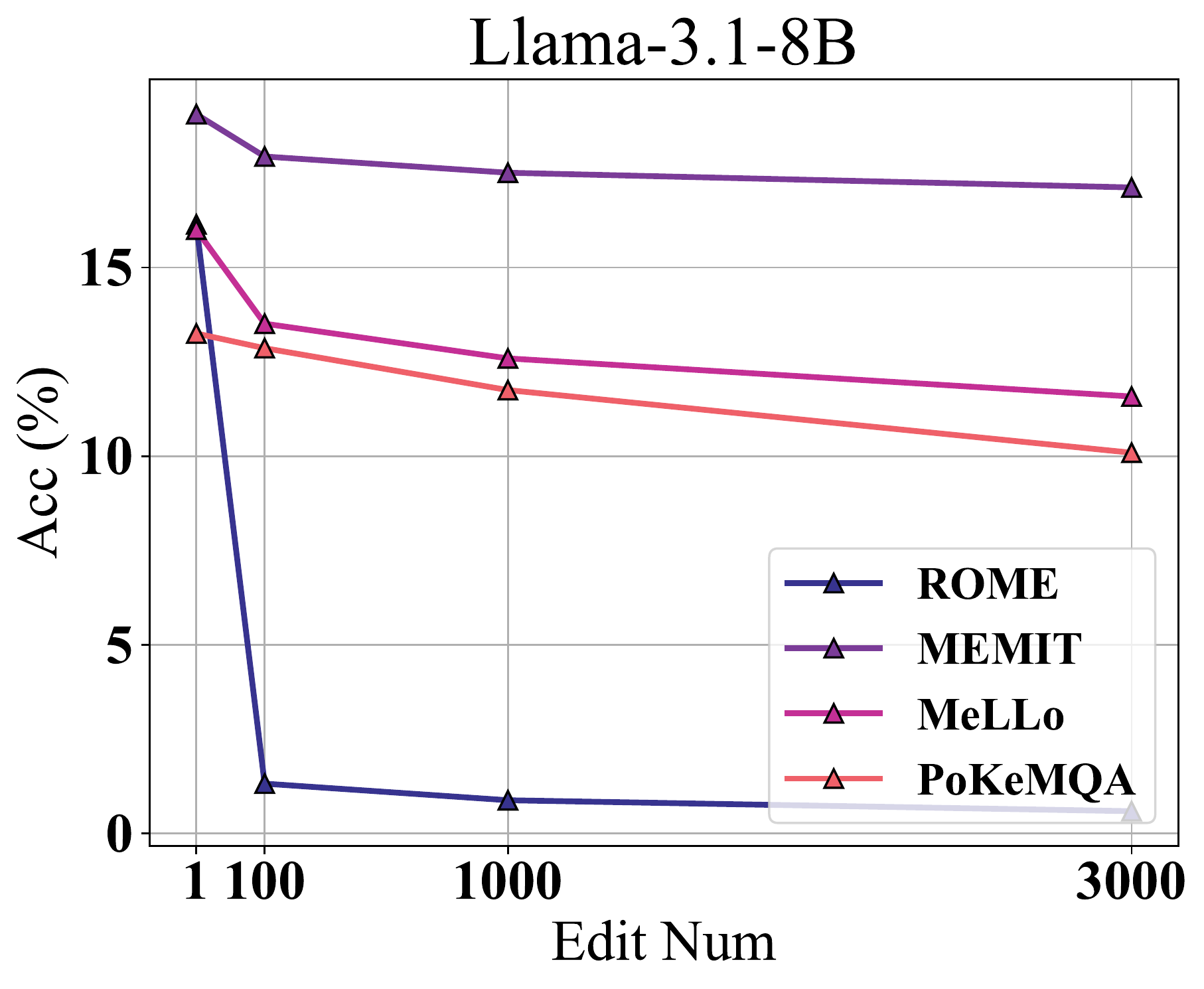}
    \end{minipage}
  \caption{Variation of \textit{Accuracy} (\texttt{Acc}) across \textsc{Qwen2.5-3B}, \textsc{Qwen2.5-7B}, 
  and \textsc{Llama-3.1-8B} models with varying edit numbers. Results for \textsc{GPT-3.5-Turbo} and 
  \textsc{GPT-4o-mini} are provided in Appendix~\ref{app:batch_extra_results}.}
  \vspace{-2.7ex}
    \label{fig:batch}
\end{figure*}
In this section, we present a thorough evaluation of state-of-the-art knowledge editing methods on \OurDATA{}. Our analysis focuses on three key aspects: the ability to recall newly added knowledge, the retention of existing knowledge, and overall accuracy. We further investigate how the performance of these methods varies with increasing edit batch sizes (\textit{i.e.}, the number of edits applied simultaneously). Through detailed case studies, we identify several noteworthy phenomena, such as overfitting in parameter-based approaches, model collapse as batch size grows, and the omission phenomenon observed in memory-based methods.

\subsection{Experimental Settings}
\noindent{\bf Language Models.}
We conduct experiments using five different target LLMs corresponding to 
three model families. 
For open source models, we select 
\textsc{LLaMa-3.1-8B-Instruct}~\citep{Dubey2024TheL3}, 
\textsc{Qwen2.5-3B-Instruct}~\citep{qwen2.5}, 
\textsc{Qwen2.5-7B-Instruct}~\citep{qwen2.5}. 
For closed source models, we select \textsc{GPT-3.5-turbo}
and \textsc{GPT-4o-mini}~\citep{Achiam2023gptGPT4TR}.

\noindent{\bf Baselines.} 
For performance comparison, we use the best performing
methods for MQA under KE as baselines. 
These include parameter-based variants: 
ROME~\citep{meng2022locating}, and
MEMIT~\citep{meng2022mass}; 
and memory-based variants:
MeLLo~\citep{zhong2023mquake}, and
PokeMQA~\citep{gu2023pokemqa}.
Since \textsc{GPT-3.5-turbo}
and \textsc{GPT-4o-mini} can only be accessed through APIs, parameter-based knowledge editing methods cannot be applied to them.

\noindent{\bf Evaluation Metrics.} We use the following metrics for evaluation: \\
\noindent \textit{(i) Augment Accuracy} (\texttt{Aug}): Measures the number of newly introduced entities that are correctly added to the answer list after the knowledge edit, relative to the original list.

\noindent \textit{(ii) Retain Accuracy} (\texttt{Ret}): Quantifies the number of entities that remain present in both the original and edited answer lists, reflecting the model's ability to retain unaltered knowledge.

\noindent \textit{(iii) Accuracy} (\texttt{Acc}): Calculated as the average of \texttt{Aug} and \texttt{Ret}, this metric provides an overall assessment of the model's capability to answer complex questions following knowledge editing.
Detailed description and mathematical formulations of these evaluation metrics are provided in Appendix~\ref{app:metric-setting}.

\paragraph{\bf Example 4.}
As illustrated in Figure~\ref{fig:datasets}, suppose the original answer is \{Los Angeles\}, and after editing, it becomes \{San Francisco, Los Angeles\}. The \texttt{Aug} metric checks if the model successfully adds the new entity (\{San Francisco\}) to its answer, while the \texttt{Ret} metric evaluates whether the model preserves the existing entity (\{Los Angeles\}) across both versions. The overall \texttt{Acc} score, computed as the average of \texttt{Aug} and \texttt{Ret}, reflects how well the model incorporates new knowledge without losing previously acquired information.

\noindent{\bf Experiment Setup.} 
We conduct experiments on varying scales of knowledge 
edits, \textit{i.e.,} using a batch of $k$-edits at a time 
with $k =\{1, 100, 1000, 3000\}$. 
To ensure a fair comparison with existing memory-based methods, we use the decomposition
examples of complex questions for MeLLo and PokeMQA, as prompts. 
Additional details on the experimental setting are provided in Appendix \ref{app:exp}.

\eat{For parameter-based methods, the settings of hyperparameters 
are provided in Appendix X \warn{ADD cite}.}

\subsection{Experimental Results}
Table~\ref{tab:main_results} presents the main experimental results. Overall, MeLLo attains the best performance in the 1-edit scenario on GPT-3.5-Turbo, achieving an \texttt{Aug} score of 49.21. Comparing the different approaches, we find that memory-based methods underperform on smaller models (e.g., \textsc{Qwen2.5-3B}), likely due to their dependence on strong instruction-following and reasoning abilities. Conversely, parameter-based methods are more suitable for smaller models, but their performance drops sharply as the number of edits in a batch increases. Below, we analyze these trends in greater detail.

\noindent{\bf Batch Editing (\#$k$-edits).}
Figure~\ref{fig:batch} shows how the accuracy of the four methods changes on \textsc{Qwen2.5-3B}, \textsc{Qwen2.5-7B}, and \textsc{Llama-3.1-8B} as the number of edits increases. Additional results for \textsc{GPT-3.5-Turbo} and \textsc{GPT-4o-mini} are provided in Appendix~\ref{app:batch_extra_results}.

Our findings indicate that memory-based methods exhibit a gradual decrease in performance as the edit batch size ($k$) grows. In contrast, parameter-based methods experience a much steeper decline, particularly when the number of edits surpasses a certain threshold. Notably, for $k \geq 100$, these models often lose coherence, resulting in inconsistent answers and irrelevant outputs, as further illustrated in Appendix Figure~\ref{fig:parafailed}.

\noindent{\bf Performance on Smaller Models.} 
We observe that for smaller language models, such as \textsc{Qwen2.5-3B}, memory-based knowledge editing methods underperform compared to parameter-based approaches. This performance gap can be explained by two main reasons:

\noindent \textit{(i) Limited Instruction-Following Ability.} Smaller models often lack the advanced instruction-following and reasoning skills required to interpret and execute complex prompts, especially those involving multi-step response planning or decomposition. As a result, when memory-based methods rely on the model to follow detailed instructions or structured plans, these models frequently fail to generate answers in the expected format or to complete all necessary reasoning steps.

\noindent \textit{(ii) Difficulty Integrating Edited Knowledge.} In the process of answering complex questions, smaller models struggle to effectively combine their internal knowledge with newly injected information from external edits. This makes it challenging for them to address sub-questions that require synthesizing both original and updated knowledge, leading to incomplete or incorrect answers.

A concrete example of this limitation is seen with the PokeMQA baseline, which depends heavily on the model's instruction-following capabilities. PokeMQA exhibits poor performance not only on \textsc{Qwen2.5-3B} but also on larger models like \textsc{LLaMa-3.1-8B} when those models' instruction-following is insufficient. This underscores the need for decomposition mechanisms that are robust to weaker instruction-following, especially in smaller models, as such mechanisms are critical for achieving strong performance in knowledge editing tasks.

\noindent{\bf Overfitting in Parameter-Based Methods.}
Interestingly, our experiments show that parameter-based methods can achieve surprisingly high accuracy on smaller models. For instance, in the Qwen2.5-3B (1-edit) scenario, MEMIT attains an accuracy of 22.43, far surpassing MeLLo's 3.83. This result is counterintuitive, as previous studies generally find that memory-based methods generalize better than parameter-based ones.

To better understand this phenomenon, we conducted a detailed case analysis and discovered that the high accuracy of parameter-based methods like MEMIT is largely due to overfitting. After the model's parameters are updated with new knowledge, the model tends to overproduce the newly injected information, outputting it in response to any related question—even when it is not contextually appropriate. This behavior artificially inflates the augmentation metric, as the model appears to recall the new knowledge very well, but in reality, it is simply repeating the edited content indiscriminately. Figure~\ref{fig:overfitting} illustrates how this overfitting leads to a higher augmentation score, highlighting a key limitation of parameter-based editing approaches on smaller models.

\noindent{\bf Omission Phenomenon.} 
We further investigate the performance of MeLLo by evaluating it with the original prompt templates provided in its official implementation. Our analysis reveals a notable issue: when decomposing complex questions, MeLLo's generated plans sometimes omit critical reasoning steps—most notably, the logical intersection step that is essential for correctly answering multi-hop questions. 

This omission occurs because the original prompt examples used to guide MeLLo's decomposition do not include cases that require conditional confirmation operations, such as logical intersections. As a result, MeLLo fails to generalize to questions that demand these reasoning patterns, and its decomposition plan skips necessary steps. 

This finding highlights a key limitation: the effectiveness of decomposition-based methods like MeLLo heavily depends on the diversity and representativeness of the prompt examples. If the prompt examples do not cover the full range of reasoning operations needed for complex questions, the model is likely to miss important steps during decomposition. Therefore, it is crucial to include prompt examples that closely resemble the structure and logic of the target questions to ensure robust generalization. 

A concrete example illustrating this omission phenomenon is provided in Appendix Table~\ref{failedcase}.

\color{black}
\noindent{\bf Comparision with other Datasets.}
\begin{figure}
    \centering
    \includegraphics[width=0.99\linewidth]{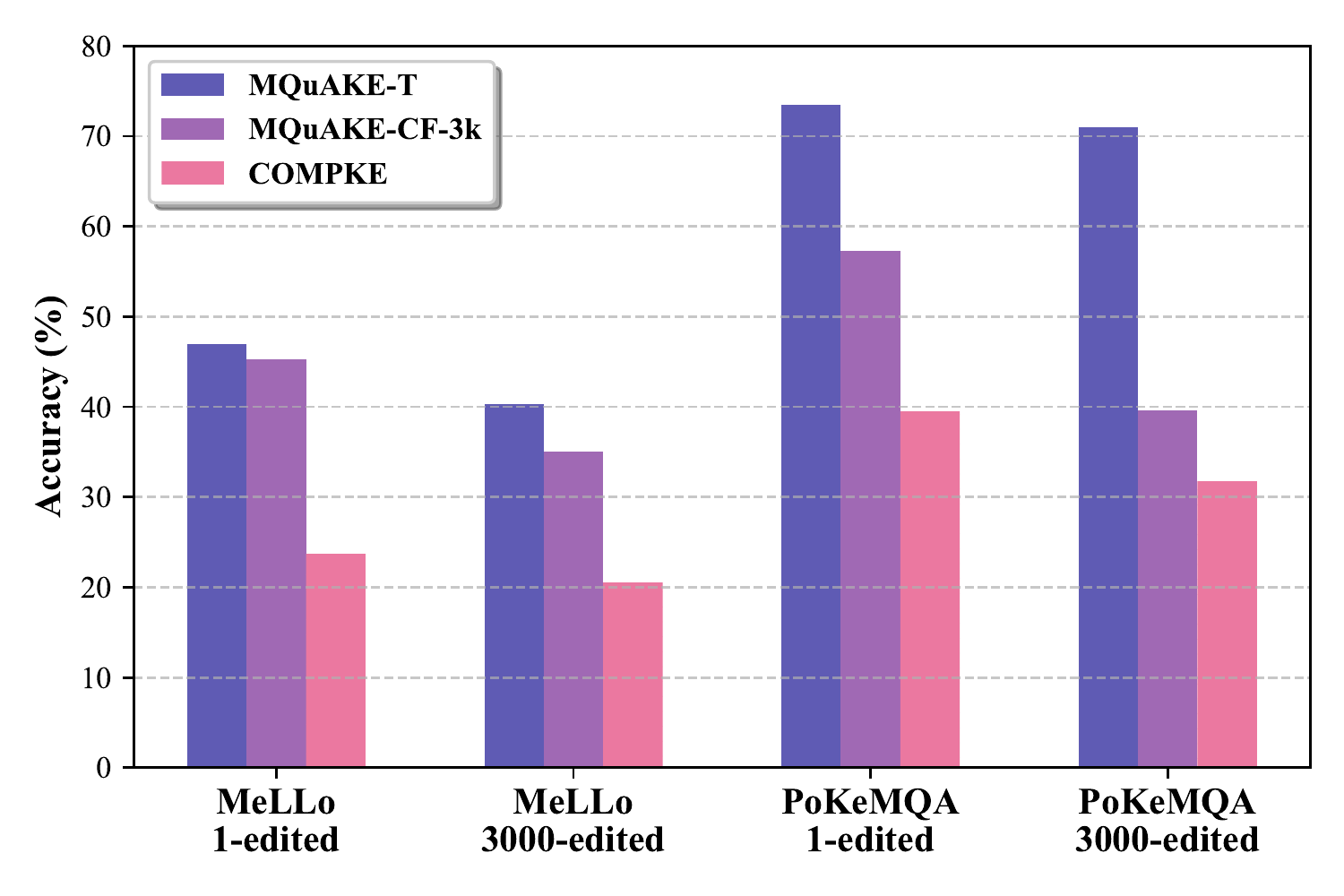}
    \caption{Performance comparison of MeLLo and PoKeMQA on the MQuAKE-T, MQuAKE-CF-3k, and \OurDATA{} datasets on \textsc{GPT-4o-mini}, with \OurDATA{} presenting more challenging than previous datasets.}
    \label{fig:dataset_comparision}
\end{figure}
To assess the relative difficulty of our benchmark, we compare it with two widely used knowledge editing datasets: MQuAKE-T and MQuAKE-CF-3k. We evaluate the performance of MeLLo and PoKeMQA on all three datasets using GPT-4o-mini as the test model. Details about the MQuAKE datasets and their evaluation metrics are provided in Appendix~\ref{app:mquake_stats} and~\ref{MQuAKE_metric}. As shown in Figure~\ref{fig:dataset_comparision}, both methods achieve noticeably lower accuracy on \OurDATA{} compared to the MQuAKE datasets. This result suggests that \OurDATA{} is more challenging and better suited for evaluating the robustness of knowledge editing methods on complex questions.


\section{Conclusion}
\label{sec:conclusion}
\vspace{-1.3ex}
In this paper, we formalize complex questions in knowledge editing—those requiring multi-step reasoning, logical composition, or integrating new and existing knowledge. To rigorously evaluate this challenging setting, we introduce \OurDATA{}, a benchmark designed to test current knowledge editing methods on such questions.

Our experiments show that both parameter-based and memory-based approaches struggle with complex questions, often losing answer accuracy due to overfitting or difficulty with multi-step reasoning, especially in smaller models. We analyze these failure modes, highlighting issues like limited instruction-following, challenges in integrating edits, and missing reasoning steps in decomposition-based methods.

By releasing \OurDATA{} and our evaluation framework, we aim to spur the development of more robust knowledge editing techniques. We hope future work will build on our findings to create methods that better handle the demands of complex question answering, improving the reliability of knowledge editing in large language models.

\vspace{-1.7ex}
\section*{Limitations}
\vspace{-1.2ex}
This work poses following limitations:
\begin{itemize}
    \setlength\itemsep{0em}
    \item In \OurDATA{}, edits are randomly introduced through counterfactual 
    modifications, which may result in discrepancies from actual/real-world modifications.  
    \item The fact triples in \OurDATA{} are restricted to one-to-one and 
    one-to-many relations, excluding many-to-many and many-to-one relationships.
\end{itemize}
\vspace{-1.7ex}
\section*{Ethics Statement}
\vspace{-1.2ex}
This work directly deals with updating the capability and/or editing 
the knowledge of large models. It has the potential for abuse, such 
as adding poisonous misinformation, malicious content, bias, etc. 
Keeping in view these concerns, we highlight that this work must not 
be used under critical settings.

\vspace{-1.7ex}
\section*{Acknowledgements}
\vspace{-1.2ex}
This work is supported in part by the funding BAS/1/1689-01-01, URF/1/4663-01-01, REI/1/5232-01-01, REI/1/5332-01-01, and URF/1/5508-01-01 from KAUST, and funding from KAUST - Center of Excellence for Generative AI, under award number 5940.

\bibliography{anthology,custom}
\bibliographystyle{acl_natbib}
\appendix
\clearpage

\section{Related Work}
\label{app:related}
\subsection{Additional Related Work}
In addition to the development of benchmarks, recent years have seen a surge of research into knowledge editing from multiple angles. One major research direction seeks to uncover the underlying mechanisms of knowledge editing methods via mechanistic interpretability~\cite{zhang2025eap,hong2024dissecting,yang2024makes,zhang2025mechanistic,hu2024hopfieldian,su2025understanding}. For example, several works investigate how knowledge is localized within model parameters and how edits propagate through the network \cite{wang2024knowledge, niu2024does, hase2024does, hase2024fundamental, ferrando2024primer, gupta2024model, yao2024knowledge,zhang2024locate,cheng2024leveraging}. Notably, \citet{hase2024does} demonstrate that causal tracing—a technique often used to identify where to intervene in a model—does not always reliably pinpoint the optimal location for editing. Other studies leverage computation graphs to analyze how knowledge edits affect the internal computations and representations of models \cite{yao2024knowledge}.

A second line of research aims to improve the effectiveness of knowledge editing in specific contexts or applications \cite{rozner2024knowledge, ma2024untyingreversalcursebidirectional, de2024llmr, huang2024can, deng2024editableextendknowledgeediting, peng2024eventlevelknowledgeediting, cai2024editingknowledge}. For instance, bidirectional relationship modeling has been introduced to address consistency issues that arise when editing knowledge in models that must reason about relationships in both directions \cite{ma2024untyingreversalcursebidirectional}. Additionally, real-time knowledge editing techniques have been proposed to enable models to adapt quickly in dynamic environments where facts may change frequently \cite{de2024llmr}.

This paper specifically investigates knowledge editing in the context of complex logical reasoning, an area that remains underexplored. Furthermore, another important research focus is on understanding and mitigating the side effects of knowledge editing. Editing a model’s knowledge can inadvertently impact unrelated facts or reasoning abilities, a phenomenon documented in several recent works \cite{hsueh-etal-2024-editing, gu2024modeleditingharmsgeneral, 10.1145/3543507.3583388, hua2024propagationpitfallsreasoning, yang2024butterflyeffectmodelediting, cohen2023evaluatingrippleeffectsknowledge, nishi2024representationshatteringtransformerssynthetic}. These studies highlight the need for careful evaluation of both the intended and unintended consequences of knowledge edits.

\subsection{Knowledge Graph Question Answering.}
\label{app:KGQA}
Several complex question answering datasets have been developed in the Knowledge Graph (KG) domain, inspired by KGs’ ability to store entity-specific information~\cite{ali2020fine,ali2021fine}.
For example, ComplexQuestions~\citep{bao2016constraint} assesses the ability of KG-based systems to answer queries involving multiple constraints. MetaQA~\citep{zhang2018variational} is a multi-hop dataset in the movie domain that includes both textual and audio modalities, requiring reasoning over up to three hops. ComplexWebQuestions~\citep{talmor2018web}, constructed on the Freebase knowledge base, features complex questions that require aggregating information from multiple web sources. CR-LT-KGQA~\citep{guo2024cr} targets commonsense reasoning and long-tail knowledge. 


While complex question answering has been widely explored in the knowledge graph (KG) community, existing KGQA datasets are not directly suitable for evaluating knowledge editing (KE) methods. This is primarily due to two fundamental limitations:

\noindent \underline{\textit{(i) Lack of explicit sub-question decomposition.}}  
Most KGQA datasets do not provide the intermediate sub-questions that compose a complex question. For instance, the ComplexQuestions dataset~\citep{bao2016constraint} contains only the overall question and its final answer, omitting any breakdown into simpler reasoning steps. Similarly, ComplexWebQuestions~\citep{talmor2018web} offers only a SPARQL query for each question, which encodes the reasoning path but does not explicitly enumerate the sub-questions. In the context of knowledge editing, it is often necessary to target and modify specific sub-components of a reasoning chain. Without clearly defined sub-questions, it becomes infeasible to perform or evaluate fine-grained edits, as there is no direct mapping between edits and the reasoning steps they affect.

\noindent \underline{\textit{(ii) Insufficient reliance on model-internal knowledge.}}  
Another key issue is that KGQA datasets typically assume access to an external knowledge base (the KG itself) for answering questions. As a result, models can answer questions by retrieving facts from the KG, rather than relying on their own parametric (internal) knowledge. In contrast, knowledge editing research focuses on modifying and evaluating the information stored within the model itself. If a dataset requires knowledge that the model has not already learned, or that is not present in its parameters, then editing operations and their evaluation become unreliable: the model may fail to answer correctly regardless of whether the edit was successful. To address this, when constructing \OurDATA{}, we carefully filter out any knowledge instances that the model cannot already recall, ensuring that all evaluated edits pertain to knowledge the model actually possesses.

In summary, the absence of explicit sub-question structure and the lack of dependence on model-internal knowledge make standard KGQA datasets ill-suited for knowledge editing research. Our dataset construction process is designed to overcome these challenges.

\section{Additional Preliminaries}
\subsection{Multi-hop Question Answering}
\label{app:mqa}
A multi-hop question can be represented as $s_1\xrightarrow{r_1} s_2 \cdots \xrightarrow{r_{n-1}} s_n$, continuously mapping one entity to another. For example. consider the question "Who is the spouse of president of U.S.", it an be represented as $\text{U.S.} \xrightarrow{\text{president is}} \text{Donald Trump} \xrightarrow{\text{spouse is}} \text{Melania Trump}$.

\subsection{Multi-hop Question Answering under KE.} 
We use $e=(s,r,o \to o^\prime)$ to represent 
a knowledge edit indicating that the object entity of subject $s$ with 
relation $r$ is updated from $o$ to $o^\prime$. This task is to solve multi-hop questions under a batch of knowledge edits $\mathcal{E}=\{e_1,e_2,\cdots\}$.

\subsection{MQA with Complex Question Answering.} 
We consider the previously studied linear 
multi-hop questions as a special case of 
complex questions involving continuous mapping 
of entity through a series of relational links, forming a one-way graph chain: 
$S_1 \overset{L_1}{\rightarrow} S_2 
\overset{L_2}{\rightarrow} \cdots \overset{L_{n-1}}{\rightarrow} S_n$, 
where $n$ represents the number of reasoning hops. 
Note that compared to complex questions,  here the intermediate set $S_i$ only encompasses a single 
entity, and $L_i$ only covers one-to-one relation mapping.

\section{\OurDATA{} (Additional Details)}
\label{app:our_data}

\begin{figure*}
    \centering
    \includegraphics[width=0.8\linewidth]{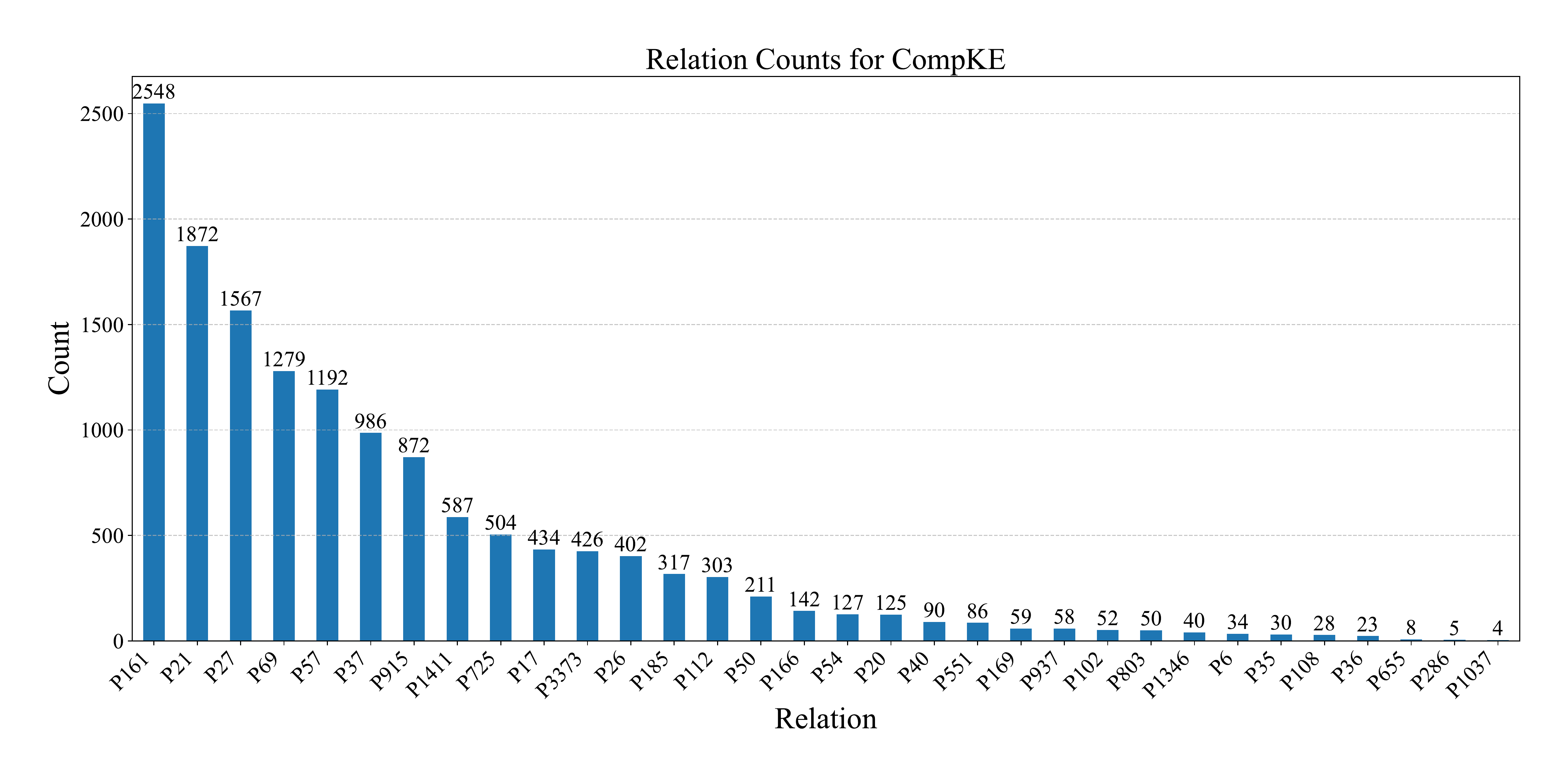}
      \vspace{-0.6cm}
    \caption{Relations and their frequencies in \OurDATA{}}
    \label{fig:relation_counts}
\end{figure*}

\begin{figure*}[!htbp]  
    \centering{
        \includegraphics[width=0.9\textwidth]{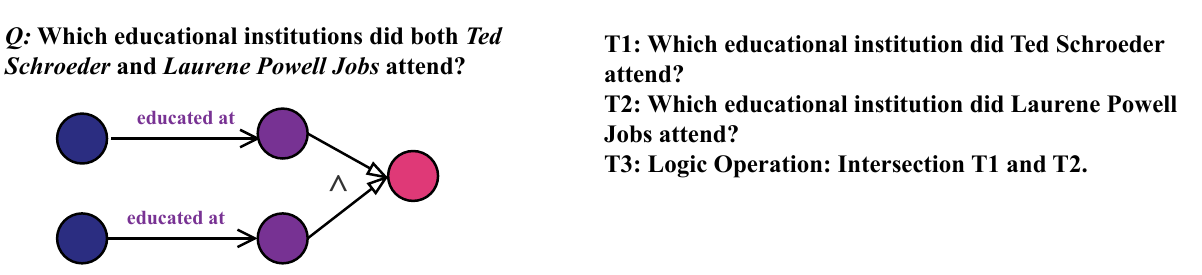}
    }{
        \includegraphics[width=0.9\textwidth]{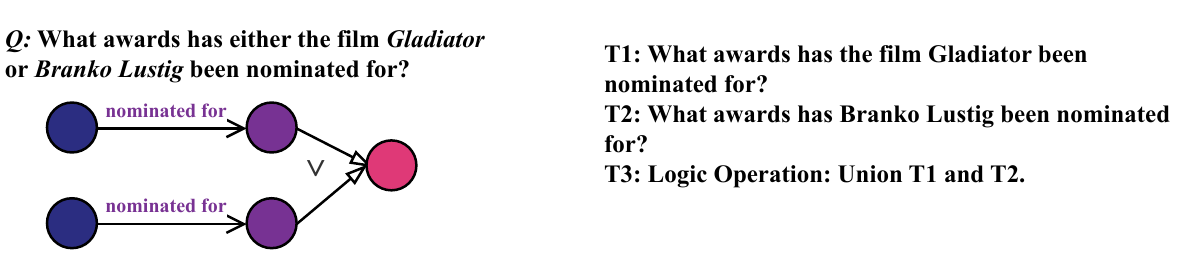}
    }{
        \includegraphics[width=0.9\textwidth]{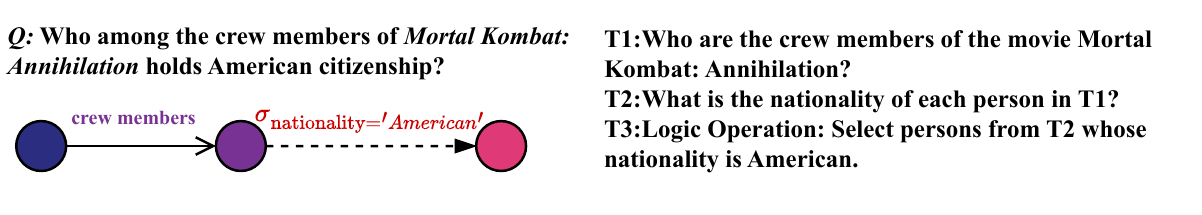}
    }{
        \includegraphics[width=0.9\textwidth]{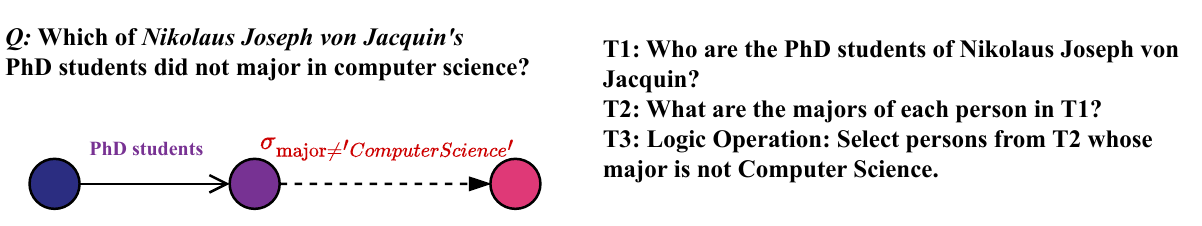}
    }{
        \includegraphics[width=0.9\textwidth]{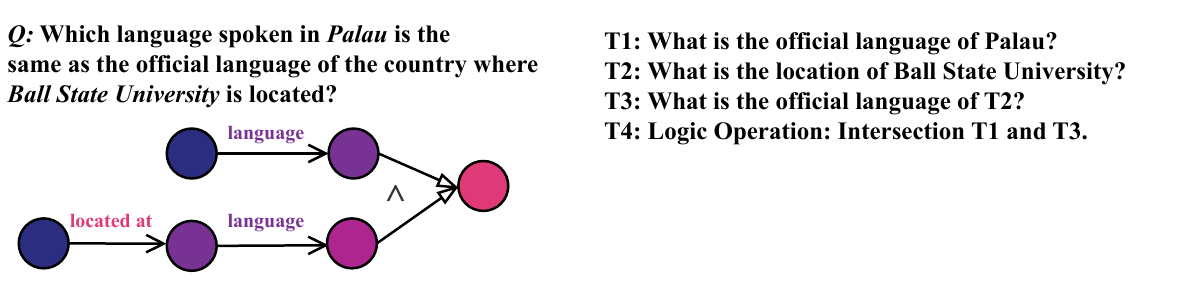}
    }{
        \includegraphics[width=0.9\textwidth]{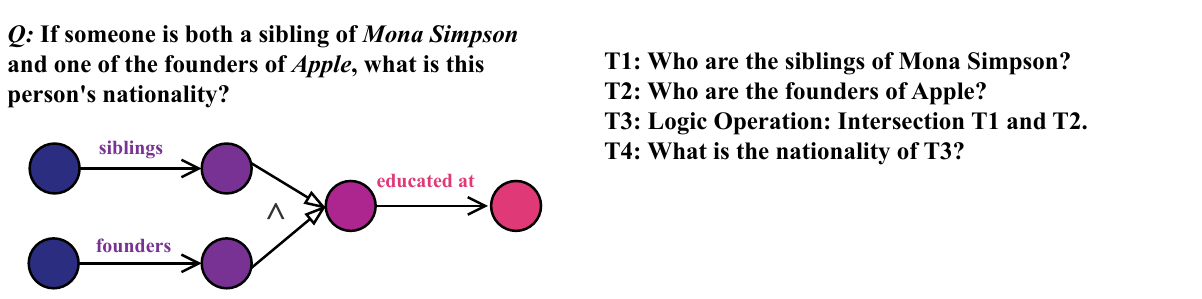}
    }
    {\includegraphics[width=0.9\textwidth]{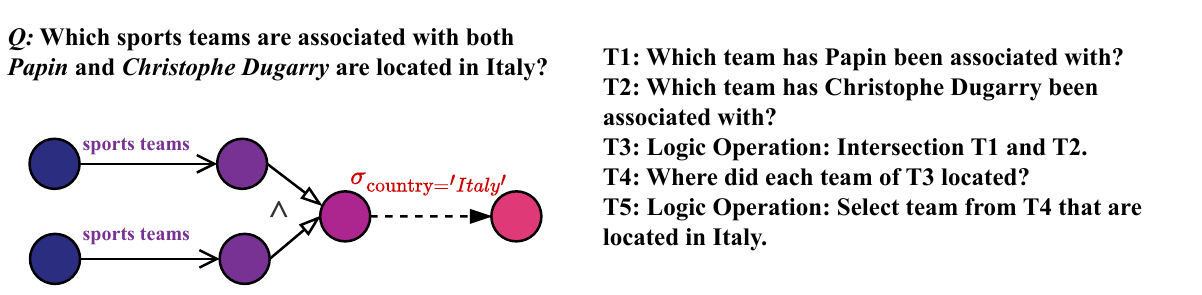}
    }
      \caption{Some typical reasoning structure in \OurDATA}
      \label{Structure}
\end{figure*}

Figure \ref{fig:dataprocess} shows the process by which we construct 
complex question. Figure \ref{Structure} gives some examples of the 
structures in \OurDATA{} and the corresponding decomposition methods. 
Table \ref{SPARQL} gives the SPARQL which we used to sample facts from WikiData. Table \ref{prompt:step6} presents the prompt used for converting structured triples into natural language. Figure~\ref{fig:relation_counts} displays the distribution of relation counts across triplets in \OurDATA{}.

\section{Additional Experimental Settings}
\label{app:exp}
\subsection{MQuAKE}
\label{app:mquake_stats}
The existing data~\textsc{MQuAKE} includes two datasets:~\textsc{MQuAKE-CF-3K}, which is based on counterfactual editing, and \textsc{MQuAKE-T}, which is based on real-world changes. These datasets cover k-hop questions ($k\in \{2,3,4\}$), each associated with one or more edits. Statistics are presented in Table~\ref{tab:mquake_dataset}.
\begin{table}[ht]
\centering
\resizebox{1.0\linewidth}{!}{
\begin{tabular}{llllll}
\toprule[1.0pt]
\textbf{Datasets} & \textbf{\#Edits} & \textbf{2-hop} & \textbf{3-hop} & \textbf{4-hop} & \textbf{Total}\\
\midrule
\multirow{5}{*}{\textbf{\textsc{MQuAKE-CF-3K}}} &1  &513  &356  &224 &1,093 \\
&2  &487  &334  &246 &1,067 \\  
&3  &-  &310  &262 &572 \\
&4  &-  &-  &268 &268 \\
&All  &1,000  &1,000  &1,000 &3,000 \\ 
\midrule
\textbf{\textsc{MQuAKE-T}} &1  &1,421  &445  &2 &1,868 \\
\bottomrule[1.0pt]
\end{tabular}}
\caption{Statistics of the \textsc{MQuAke} dataset.}
\label{tab:mquake_dataset}
\end{table}

\subsection{Baselines}
\label{app:exp_baselines}
\noindent {\bf ROME.} ROME by~\citet{meng2022locating} uses a locate-then-edit 
paradigm. For a specific knowledge editing, ROME employs causal tracing to 
pin-point the exact layer of the MLP module within the Transformer model 
architecture that encodes the paticular factual association. Then it will 
perform a rank-one modification on the identified layer.

\noindent {\bf MEMIT.} MEMIT by~\citet{meng2022mass} is an evolution of ROME 
to transcend the inherent limitation that ROME can only edit a single fact at 
a time. At a time, MEMIT can identify and modify multiple layers in a single 
pass, allowing for the simultaneous editing of numerous facts.

\noindent {\bf MeLLo.} MeLLo by~\citet{zhong2023mquake} adopts a strategy that 
alternates between planning and solving stage to solve multi-hop question. It 
employ a semantic-based retrieval to retrieve relevant edits, and a self-checking 
mechanism to enable the model to assess the relevance of edits and modifications.

\noindent {\bf PokeMQA.} PokeMQA by~\citet{gu2023pokemqa} is 
a memory-based method that extends MeLLo and proposes a 
two-stage retrieval process to enhance the success rate of retrieving relevant edits.

\begin{figure*}[ht]
    \centering
    \begin{minipage}{0.45\linewidth}
        \centering
        \includegraphics[width=\linewidth]{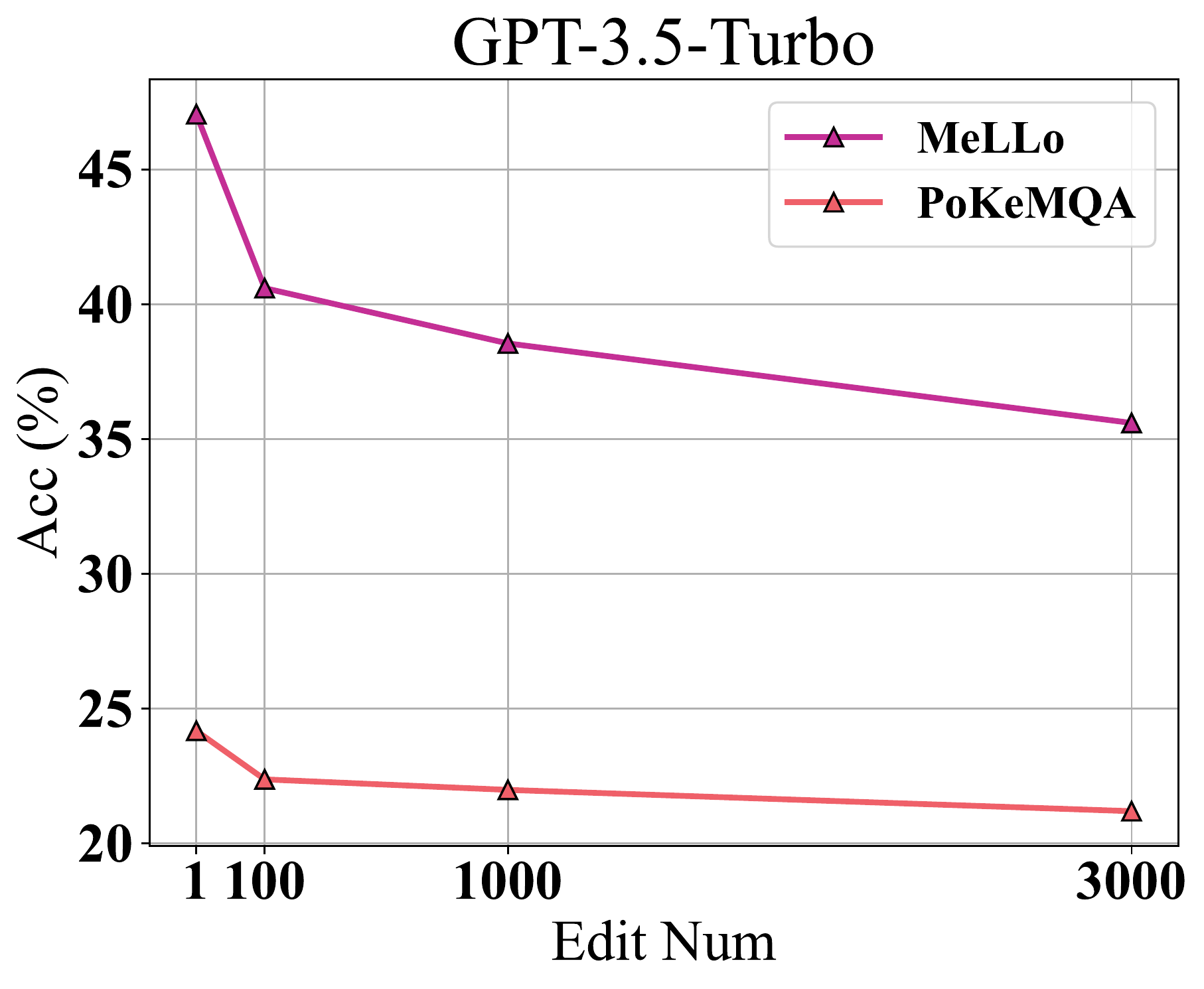}
    \end{minipage} \hfill
    \begin{minipage}{0.45\linewidth}
        \centering
        \includegraphics[width=\linewidth]{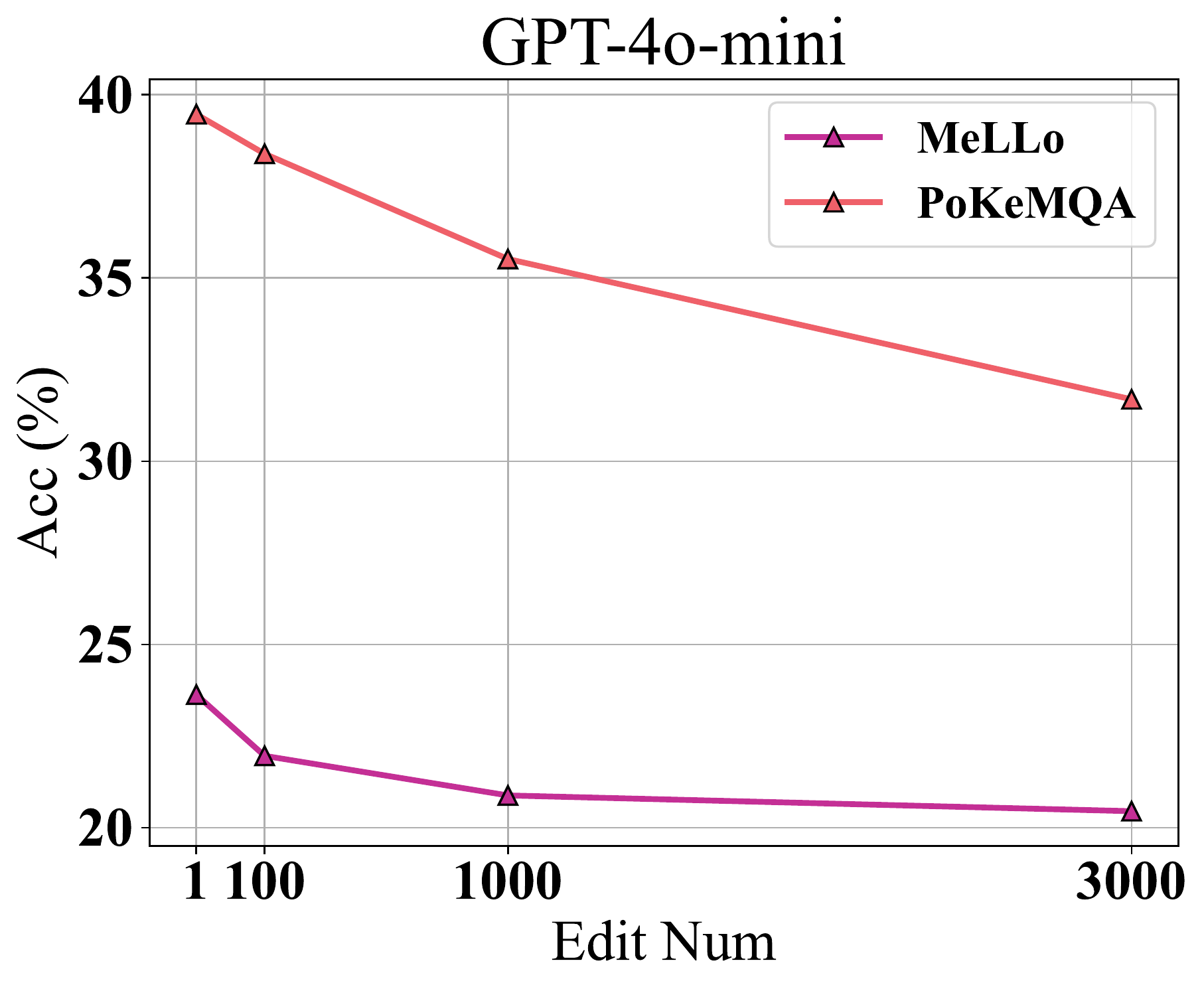}
    \end{minipage}
        \caption{{Variation of \textit{Accuracy (Acc)} across  GPT-3.5-Turbo and GPT-4o-mini models with varying edit numbers.}}
    \label{fig:batch_app}
\end{figure*}
\subsection{Evaluation Metrics}
\label{app:metric-setting}
Detailed metrics and mathematical definitions are given below:

\noindent \textbf{(i) Augment Accuracy (Aug)} is used to measure whether the 
edited model can response added knowledge on complex questions. The formula 
for calculating Aug-Acc is as follows:
\begin{equation}
\mathbb{E}_{q\in \mathcal{Q}} (\left| M^\prime(q) \cap \mathcal{A}_{aug}\right| / \left|\mathcal{A}_{aug}\right|)
\end{equation}
Where $M^\prime(\cdot)$ represents the edited model, and $\mathcal{Q}$ denote 
the datasets for complex questions, $\mathcal{A}_{aug}=\mathcal{A}^\prime \setminus \mathcal{A}$, $\mathcal{A}^\prime$ is edited answer set and $\mathcal{A}$ is original answer set.

\noindent \textbf{(ii) Retention Accuracy (Ret)} is used to measure whether the 
edited model can retain the original knowledge on complex questions. The formula 
for calculating Ret-Acc is as follows:
\begin{equation}
\mathbb{E}_{q\in \mathcal{Q}} (\left| M^\prime(q) \cap \mathcal{A}_{ret}\right| / \left|\mathcal{A}_{ret}\right|)
\end{equation}
Where $\mathcal{A}_{ret}=\mathcal{A}^\prime \cap \mathcal{A}$.

\label{MQuAKE_metric}
\noindent \textbf{(iii) Multi-hop Accuracy (M-Acc)} is used to measure the accuracy 
for multi-hop question under knowledge editing(i.e.,MQuAKE). The formula for calculating M-Acc is as follows:
\begin{equation}
\mathbbm{1}\left[\bigvee_{q \in \mathcal{Q}} [M^\prime(q) = a^\prime]\right]. 
\end{equation}
Where $M^\prime(\cdot)$ represents the edited model, and $\mathcal{Q}$ and $a^\prime$ 
denote the multi-hop questions and the final-hop answers for each data, respectively.

\subsection{Experiment Setup}
Table \ref{hyper} shows the hyperparameter settings for the parameter-based 
methods. For the experiments involving ROME and MEMIT, we utilized 
four NVIDIA Tesla L20 GPUs, with 48GB of memory. A single 
RTX 4090 GPU was used for MeLLo and PokeMQA.

\begin{table*}[ht]
    \centering
    \small
    \noindent\fbox{%
    \begin{minipage}{2.0\columnwidth} 
    \tt 
 \textbf{ROME:}\\
 layers: [5], \\
 fact\_token: subject\_last, \\
 v\_num\_grad\_steps: 25(for Llama-3.1-8B)||15(for Qwen2.5), \\
 v\_lr: 5e-1, \\
 v\_loss\_layer: 31(for Llama-3.1-8B)||27(for Qwen2.5-7B)||35(for Qwen2.5-3B), \\
 v\_weight\_decay: 1e-3, \\
 clamp\_norm\_factor: 4, \\
 kl\_factor: 0.0625, \\
 mom2\_adjustment: false, \\
 context\_template\_length\_params: [[5, 10], [10, 10]]\\

\textbf{MEMIT:}\\
    layers: [3,4,5,6,7,8],\\
    clamp\_norm\_factor: 4,\\
    layer\_selection: all,\\
    fact\_token: subject\_last,\\
    v\_num\_grad\_steps: 25(for Llama-3.1-8B)||15(for Qwen2.5), \\
    v\_lr: 5e-1,\\
    v\_loss\_layer: 31(for Llama-3.1-8B)||27(for Qwen2.5-7B)||35(for Qwen2.5-3B), \\
    v\_weight\_decay: 1e-3,\\
    kl\_factor: 0.0625,\\
    mom2\_adjustment: true,\\
    mom2\_update\_weight: 15000,\\
    mom2\_dataset: wikipedia,\\
    mom2\_n\_samples: 100000,\\
    mom2\_dtype: float32\\
    \end{minipage}
    }
    \caption{Several key hyperparameters for parameter-based KE methods}
    \label{hyper}
\end{table*}

\section{Additional Experimental results}
\subsection{An example for overfitting phenomenon of parameter-based methods.}
\definecolor{c1}{RGB}{0, 102, 51}

\begin{figure*}
\centering    \includegraphics[width=0.99\linewidth]{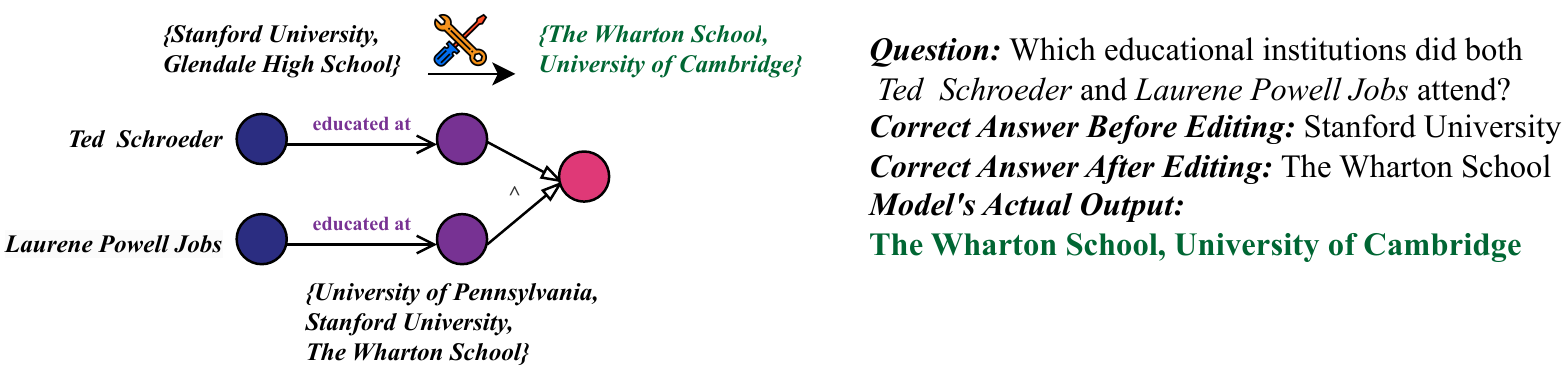}
    \caption{An example of MEMIT applied to Qwen2.5-3B. The correct solution should be for the model to take the \textbf{intersection} of the new knowledge about Ted Schroeder's schools \textbf{\textit{\{The Wharton School, University of Cambridge\}}} and Laurene Powell Jobs' schools \textbf{\textit{\{University of Pennsylvania, Stanford University, The Wharton School\}}}, yielding the final answer: \textbf{\textit{The Wharton School}}. However, the model's output is \textbf{\textit{\textcolor{c1}{\{The Wharton School, University of Cambridge\}}}} (i.e., the newly injected knowledge). While the correct entity \textit{The Wharton School} is correctly included in the final answer(i.e., metric \texttt{Aug} is judged as correct), the model after editing merely reiterated the injected knowledge \textbf{without} performing the intersection operation. This example illustrates why the \texttt{Aug} metric is overestimated.}
    \label{fig:overfitting}
\end{figure*}

\label{app:overfitting}
Figure \ref{fig:overfitting} shows an example of overfitting phenomenon when MEMIT is applied to Qwen2.5-3B.
\subsection{Results for Batch Editing(\#$k$-edits)}
\label{app:batch_extra_results}
The results of GPT-3.5-Turbo and GPT-4o-mini for the batch editing, \textit{i.e.,} varying the number 
of edits ($k$) are presented in Figure~\ref{fig:batch_app}.

\begin{table*}
    \centering
    \small
    \noindent\fbox{%
    \begin{minipage}{2.0\columnwidth} 
    \tt
\textbf{User Input:} s1="Electric Dreams",r1="the crew members are",s2="Michael Madsen",r2="siblings are"\\
\textbf{System Output:}\\
Who is someone that is both a member of the cast of the movie Electric Dreams and a sibling of Michael Madsen?\\
Of all the cast members of the film Electric Dreams, who are also Michael Madsen's siblings?\\
Which members of the cast of the film Electric Dreams are also siblings of Michael Madsen?\\
\\
\textbf{User Input:} s1="Felix Sanchez",r1="educated at",s2="Maria Sanchez",r2="educated at"\\
\textbf{System Output:}\\
Which schools have Maria Sanchez and Felix Sanchez both attended?\\
In the places where Maria Sanchez and Felix Sanchez received their education, which schools do they have in common?\\
Which educational institutions did Maria Sanchez and Felix Sanchez both go to?\\
\\
\textbf{User Input:} s1="Benoit Jacquot",r1="awards that subject was nominated for",s2="The Public Woman",r2="awards that subject was nominated for"\\
\textbf{System Output:}\\
What awards have Benoit Jacquot and "The Public Woman" both been nominated for?\\
What nominations do Benoit Jacquot and "The Public Woman" share for awards?\\
What award nominations are shared by Benoit Jacquot and "The Public Woman"?\\
\\
\textbf{User Input:} s1="Clara Agnelli",r1="siblings are",s2="Giovanni Agnelli",r2="The founders of the company are"  \\
\textbf{System Output:}\\
Who are both siblings of Clara Agnelli and founders of the company Giovanni Agnelli started?\\
Among Clara Agnelli's siblings, who are also recognized as the founders of the company by Giovanni Agnelli?\\
Which individuals are both siblings of Clara Agnelli and acknowledged as the founders of the company by Giovanni Agnelli?\\
    \end{minipage}
    }
    \caption{A prompt used for converting structured triples into natural language.}
    \label{prompt:step6}
\end{table*}

\begin{table*}[ht]
\centering
\captionsetup{justification=centering}  
\begin{tabular}{|c|c|}
\hline
\textbf{SQL Query} & \textbf{Description} \\
\hline
\begin{minipage}[t]{0.45\textwidth}
\begin{verbatim}
SELECT ?object WHERE {
  wd:{qid} wdt:pid ?object.
  FILTER(LANG(?object) = "en").
}

\end{verbatim}
\end{minipage} &
\begin{minipage}[t]{0.45\textwidth}
This SPARQL query retrieves the object associated with the <pid> of entity. 
\end{minipage} \\
\hline
\begin{minipage}[t]{0.45\textwidth}
\begin{verbatim}
SELECT (COUNT(?statement) AS 
?referencesCount) WHERE {
  wd:{entity_id} ?p ?statement.
  ?statement 
  prov:wasDerivedFrom ?source.
}

\end{verbatim}
\end{minipage} &
\begin{minipage}[t]{0.45\textwidth}
This SPARQL query retrieves the count of references (i.e., the number of statements that refer to a source) for a specific entity. This query is used to filters out triples with low references counts(i.e.,unpopular entity).
\end{minipage} \\
\hline
\begin{minipage}[t]{0.45\textwidth}
\begin{verbatim}
SELECT ?alias WHERE {
  wd:{qid} skos:altLabel ?alias.
  FILTER(LANG(?alias) = "en").
}

\end{verbatim}
\end{minipage} &
\begin{minipage}[t]{0.45\textwidth}
This SPARQL query retrieves the aliases associated with the entity, 
\end{minipage} \\
\hline
\end{tabular}
\caption{SPARQL Queries and Descriptions}
\label{SPARQL}
\end{table*}
\begin{figure*}[ht]
    \centering
\includegraphics[width=0.8\linewidth]{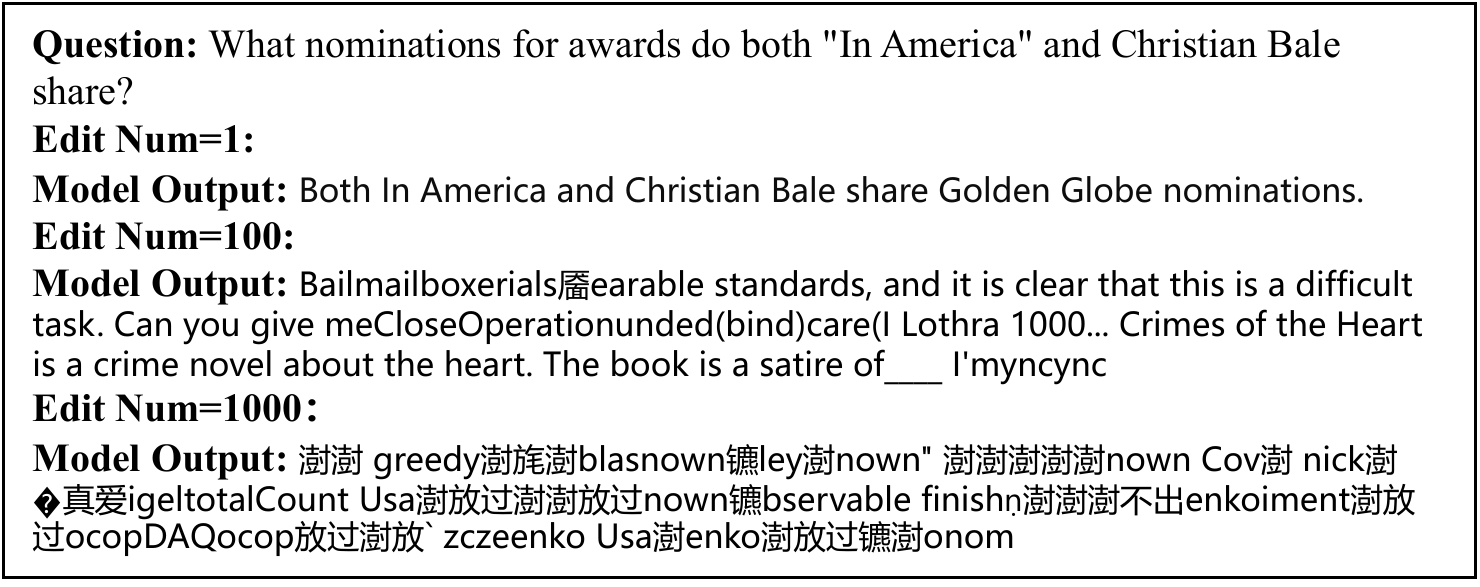}
    \caption{When the edit batch size increases, the MEMIT method outputs a large amount of gibberish after models like Qwen-2.5-3B and other smaller models.}
    \label{fig:parafailed}
\end{figure*}

\definecolor{maskcolor}{HTML}{6e9feb}
\definecolor{c2}{RGB}{150,120,255} 
\begin{table*}[ht]
    \centering
    \small
    \noindent\fbox{%
    \begin{minipage}{2.0\columnwidth} 
    \tt 
    
    Question: \textbf{If a person is the director of both Thief and Ali, which country does this person belong to?}\\
\textbf{Subquestion:} Who is the director of Thief? \\ 
\textbf{Generated answer:} The director of Thief is Michael Mann.  
\\
\textbf{Retrieval:}The country to which Michael Mann belongs is Italy
\\
Retrieved fact does not contradict to generated answer, so the intermediate answer is: Italy.\\
\textbf{Subquestion:} Who is the director of Ali?  \\
\textbf{Generated answer:} The director of Ali is Michael Mann.  
\\
\textbf{Retrieval:}The country to which Michael Mann belongs is Italy
\\
Retrieved fact does not contradict to generated answer, so the intermediate answer is: Italy.  \\
\textbf{Final answer:} Italy\\
    \end{minipage}
    }

    \caption{The decomposition of a complex question by Mello did not take into account logical operations.}
    \label{failedcase}
\end{table*}

\begin{table*}[t]
    \resizebox{0.95\linewidth}{!}{
    \begin{tabular}{l|l|l}
    \toprule 
     Relation & Question template & Cloze-style statement template \\
     \midrule
P40 & Who are [S]'s children? & [S]'s children are \\
P69 & Where did [S] receive education? & The university where [S] was educated is \\
P3373 & Who are the siblings of [S]? & [S]'s siblings are \\
P50 & Who are the author(s) of [S]? (list all) & The author(s) of [S] is(are) \\
P161 & Who are the cast members of movie [S]? & The cast members of movie [S] are \\
P112 & Who are the people who founded company [S]? & The people who founded Company [S] are \\
P54 & Which organizations is [S] a member of? & [S] is a member of the following organizations \\
P915 & Where were movie [S] filmed? & The movie [S] was filmed at \\
P37 & What are the official languages of country [S]? & The official languages of country [S] are \\
P1830 & Which companies does S own? & [S] owns the following companies \\
P6 & Who are the heads of government for [S]? & The heads of government for [S] are \\
P803 & What are the professorship ranks for [S]? & The professorship ranks for [S] are \\
P185 & Who are the doctoral students of [S]? & The doctoral students of [S] are \\
P57 & Who is the director of the film [S]? & The film [S] is directed by \\
P1411 & What awards was the film [S] nominated for? & The film [S] is nominated for \\
P1346 & Who are the winners for [S] prize? & The winners for [S] prize are \\
P286 & Who are the head coaches for team [S]? & The head coaches for team [S] are \\
P166 & What awards did [S] receive? & The award received by [S] are \\
P800 & What are the notable works of [S]? & The notable works of [S] are \\
P725 & Who are the voice actors in the movie [S]? & The voice actor in the movie [S] are \\
P655 & Who are the translators of the book [S]? & The translators of the book [S] are \\
P27 & Which country is [S] a citizen of? & The country to which [S] belongs is \\
P21 & What's [S]'s gender? & [S]'s gender is \\
P169 & Who is the CEO of company [S]? & The CEO of company [S] is \\
P35 & Who is the head of state of country [S]? & The head of state of country [S] is \\
P26 & Who is the spouse of [S]? & The spouse of [S] is \\
P1037 & Who is the director of [S]? & The director of [S] is \\
P20 & In which city did [S] die? & [S] died in the city of \\
P551 & Where does [S] live? & [S] lives in the place of \\
P159 & Where is the headquarters of company [S]? & The headquarters of company [S] is located in \\
P17 & In which country is [S] located? & [S] is located in the country of \\
P108 & Who is the employer of [S]? & [S] is an employee in the organization of \\
P102 & Which political party is [S] affiliated with? & [S] is affiliated with the political party of \\
P937 & Where does [S] work? & [S] works in the place of \\
P140 & What is the religion of [S]? & [S] is affiliated with the religion of \\
P106 & What is [S]'s occupation? & [S]'s occupation is \\
P30 & On which continent is country [S] located? & Country [S] is located in the continent of \\
P38 & What is the currency of country [S]? & The currency of country [S] is \\
P641 & Which sport is [S] associated with? & [S] is associated with the sport of \\
P36 & What is the capital of country [S]? & The capital of country [S] is \\
\bottomrule
    \end{tabular}}
    \caption{Relations we use to construct \OurDATA}
    \label{Relation}
\end{table*}

\end{document}